\documentclass{article}

\usepackage{microtype}
\usepackage{graphicx}
\usepackage{subfig}
\usepackage{booktabs} % for professional tables
\usepackage{caption}
\usepackage{amsmath,amsfonts,bm}

\def\eqref#1{equation~\ref{#1}}
\def\1{\bm{1}}

\def\vmu{{\bm{\mu}}}
\def\vtheta{{\bm{\theta}}}
\def\vphi{{\bm{\phi}}}
\def\vpsi{{\bm{\psi}}}
\def\va{{\bm{a}}}

\def\vu{{\bm{u}}}
\def\vv{{\bm{v}}}
\def\vw{{\bm{w}}}

\DeclareMathAlphabet{\mathsfit}{\encodingdefault}{\sfdefault}{m}{sl}
\SetMathAlphabet{\mathsfit}{bold}{\encodingdefault}{\sfdefault}{bx}{n}
\newcommand{\tens}[1]{\bm{\mathsfit{#1}}}

\def\tH{{\tens{H}}}

\def\tW{{\tens{W}}}
\def\tX{{\tens{X}}}

\def\sD{{\mathbb{D}}}
\def\sS{{\mathbb{S}}}

\newtheorem{definition}{Definition}
\newtheorem{theorem}{Theorem}

\usepackage{mathtools}
\newcommand{\defines}{\coloneqq}

\DeclareMathOperator*{\argmin}{arg\,min}

\newcommand{\vspacecaption}{\vspace{-6pt}}
\newcommand{\vspacecaptionlow}{\vspace{-3pt}}

\newcommand{\vspaceequation}{\vspace{-4pt}}

\usepackage{hyperref}

\usepackage[accepted]{icml2021}

\icmltitlerunning{Robustness to Pruning Predicts Generalization in Deep Neural Networks}

\begin{document}

\twocolumn[
\icmltitle{Robustness to Pruning Predicts\\
           Generalization in Deep Neural Networks}



\icmlsetsymbol{equal}{*}

\begin{icmlauthorlist}
\icmlauthor{Lorenz Kuhn}{equal,eth}
\icmlauthor{Clare Lyle}{equal,oxford}
\icmlauthor{Aidan N. Gomez}{oxford}
\icmlauthor{Jonas Rothfuss}{eth}
\icmlauthor{Yarin Gal}{oxford}
\end{icmlauthorlist}

\icmlaffiliation{eth}{Department of Computer Science, ETH Zurich}
\icmlaffiliation{oxford}{OATML, Department of Computer Science, University of Oxford}

\icmlcorrespondingauthor{Lorenz Kuhn}{kuhnl@ethz.ch}

\icmlkeywords{Machine Learning, ICML}

\vskip 0.3in
]



\printAffiliationsAndNotice{\icmlEqualContribution}  % leave blank if no need to mention equal contribution

\begin{abstract}
     Existing generalization measures that aim to capture a model's simplicity based on parameter counts or norms fail to explain generalization in overparameterized deep neural networks. In this paper, we introduce a new, theoretically motivated measure of a network’s simplicity which we call prunability: the smallest \emph{fraction} of the network’s parameters that can be kept while pruning without adversely affecting its training loss. We show that this measure is highly predictive of a model’s generalization performance across a large set of convolutional networks trained on CIFAR-10, does not grow with network size unlike existing pruning-based measures, and exhibits high correlation with test set loss even in a particularly challenging double descent setting. Lastly, we show that the success of prunability cannot be explained by its relation to known complexity measures based on models’ margin, flatness of minima and optimization speed, finding that our new measure is similar to -- but more predictive than -- existing flatness-based measures, and that its predictions exhibit low mutual information with those of other baselines.
\end{abstract}

\section{Introduction}
\label{section_introduction}
The gap between learning-theoretic generalization bounds for highly overparameterized neural networks and their empirical generalization performance remains a fundamental mystery to the field
\citep{zhang2016understanding, jiang2019fantastic, allen2019learning}. While these models are already successfully deployed in many applications, improving our understanding of how neural networks perform on unseen data is crucial for safety-critical use cases. By understanding which factors drive generalization in neural networks we may further be able to develop more efficient and performant network architectures and training methods. 

\begin{figure}[t]
%\vskip 0.2in
\begin{center}
\centerline{\includegraphics[width=\columnwidth]{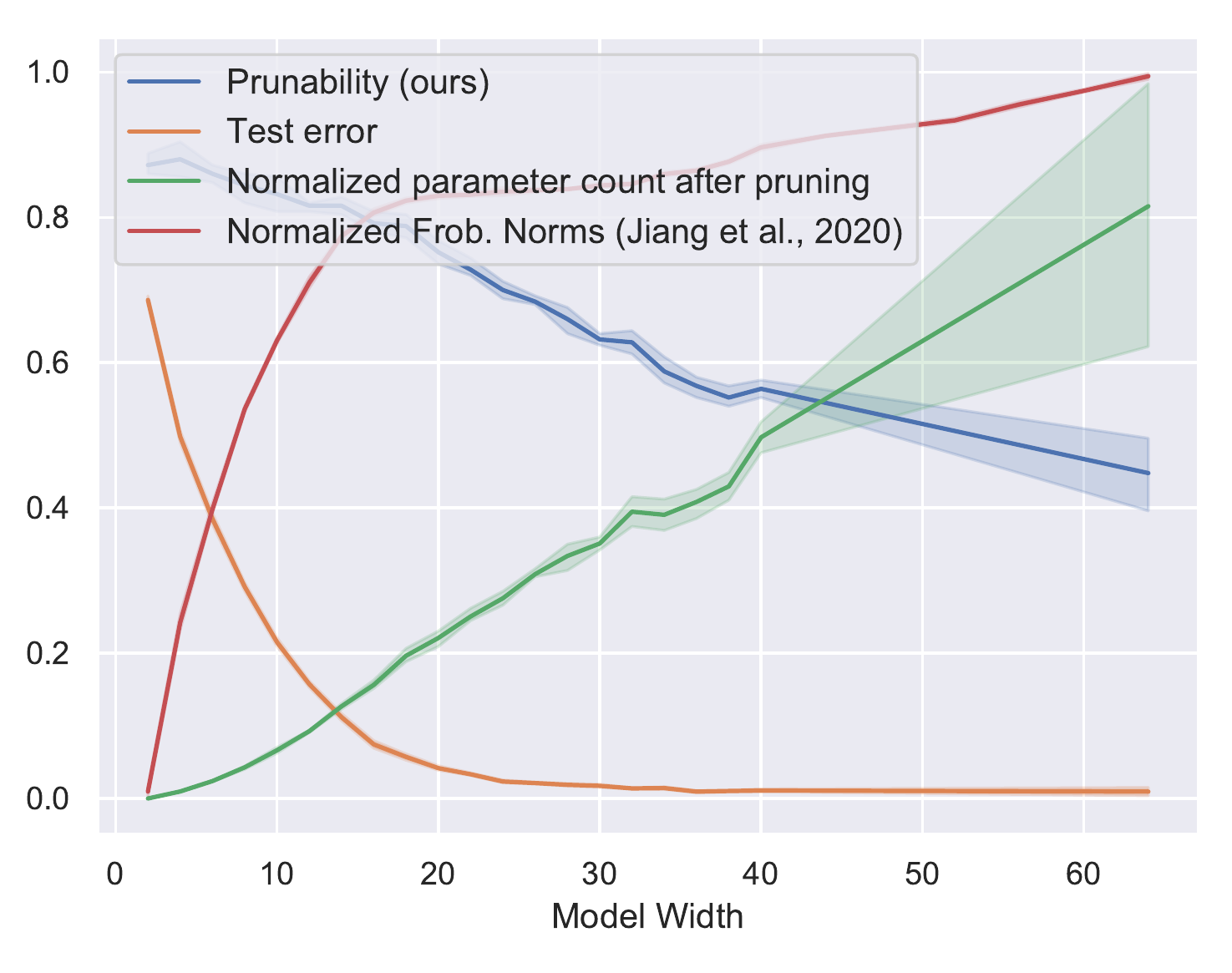}} \vspace{-6pt} 
\caption{\textbf{Existing model size-based complexity measures fail to capture the observation that lager models generalize better.} Such measures based on norms or parameter-counts grow as model size is increased and the generalization error (orange) decreases. The same applies to existing measures based on the model size of \textit{compressed} models (green) and a particularly strong norm-based baseline (red) studied by \citet{jiang2019fantastic}. We propose a new generalization measure based on the relative model size after pruning (blue) and show that this measure correctly predicts the decrease in generalization error as model size increases. This example of ResNets trained on CIFAR-10 is merely an illustration of our claim. We provide a rigorous examination in Section \ref{section_experimental_results}. \vspace{-12pt} }
\label{figure_pruned_parameter_count_vs_prunability}
\end{center}
\vskip -0.2in

\end{figure}

Numerous theoretically and empirically motivated attempts have been made to identify \textit{generalization measures}, properties of the trained model, training procedure and training data that distinguish models that generalize well from those that do not \citep{jiang2019fantastic}. A number of generalization measures have attempted to quantify Occam's razor, i.e.\ the principle that \textit{simpler} models generalize better than complex ones  \citep{neyshabur2015normbased, bartlett2017spectrally}. This has proven to be non-trivial, as many measures, particularly norm-based measures, grow with the size of the model and thus incorrectly predict that larger networks will generalize worse than smaller networks.

In this paper, we propose a new way of quantifying a model's simplicity that avoids this failure mode. We leverage the empirical observation that large fractions of trained neural networks' parameters can be \textit{pruned} -- that is, set to 0 -- without hurting the models' performance \citep{gale2019state, zhu2017prune, han2015learning}. Based on this insight, we introduce a new measure of a model's simplicity which we call \emph{prunability}: the smallest fraction of weights we can keep while pruning a network without hurting its training loss.

In a range of empirical studies, we demonstrate that a model's prunability is highly informative of its generalization. In particular, we find that the larger the fraction of parameters that can be pruned without hurting a model's training loss, the better the model will generalize.
Overall, we show that the smaller the fraction of parameters a model actually ``uses" --  the \textit{simpler} it is -- the better the network's generalization performance. 

We illustrate the limitations of existing generalization measures based on parameter counts, norms and pruning in \ref{figure_pruned_parameter_count_vs_prunability}. We compare ResNets of varying widths that are trained on CIFAR-10. We remove the smallest parameters from each model up to the point where further pruning would cause a spike of the training loss. We observe that the absolute number of parameters that remain after pruning grows with the model width, while the test error decreases with model width. A strong norm-based baseline studied by \citet{jiang2019fantastic} also grows with the model size. The fraction of parameters that remains after pruning – the prunability – decreases with the model width, correctly predicting that wider models generalize better.

Recent empirical studies \citep{jiang2019fantastic, jiang2018predicting} identify three classes of generalization measures that do seem predictive of generalization: measures that estimate the \textit{flatness} of local minima, the speed of the optimization, and the margin of training samples to decision boundaries. We empirically compare our proposed generalization measure against strong baselines from each of these classes. We find that prunability seems competitive with most of these strong baselines and that our measure seems to capture a different aspect of generalization than the existing measures. Additionally, we study a particularly challenging setting for many generalization measures – Deep Double Descent \citep{nakkiran2019deep} – and show that prunability captures this phenomenon better than previously proposed baselines.

In summary, our main contributions are thus the following:
\begin{enumerate}
    \item We introduce a new generalization measure called \textit{prunability} that captures a model's simplicity (Sec. \ref{section_prunability}) and show that across a large set of models this measure is highly informative of a network's generalization performance (Sec. \ref{section_performance_of_prunability}). While we also provide some theoretical motivation for this measure, we see our empirical study as the main contribution of this paper and an important stepping stone towards a more comprehensive theoretical understanding of prunability.
    
    \item We show that even in a particularly challenging setting in which we observe a test loss double descent \citep{nakkiran2019deep, he2016identity}, prunability is informative of models' test performance and is competitive with an existing strong generalization measures (Sec. \ref{section_double_descent_results}).
    
    \item Lastly, we investigate whether the success of prunability can be explained by its relationship to flat local minima. We find that while prunability makes similar predictions to some existing measures that estimate the flatness of minima, it differs from them in important ways, in particular exhibiting a stronger correlation with generalization performance than these flatness-based measures (Sec. \ref{section_performance_of_prunability} and \ref{section_comparison_of_generalization_measures}).
\end{enumerate}

\section{Related Work}
\label{section_related_work}
\vspacecaptionlow

\citet{rissanen1986stochastic} formalizes Occam's razor as a model's Minimum Description length, a concept which was later applied to small neural networks by \citet{hinton1993keeping}. Similarly, \citet{neyshabur2015normbased, neyshabur2017exploring} and numerous other approaches suggest that networks with smaller parameter norms generalize better. These measures, however, incorrectly predict that \textit{smaller} models generalize better than larger models, the opposite of what is observed empirically \cite{jiang2019fantastic}. \citet{zhou2018non} formulate a generalization bound that is non-vacuous even for large network architectures based on the description length of a compressed (pruned, quantized and efficiently encoded) network. While this is an elegant approach, the authors acknowledge that the compressed model size also grows with the size of the original network and thus suffers from the same problem as the other approaches based on model size described above. In contrast, we propose and evaluate a pruning-based generalization measure that is able to overcome this limitation.
 See Figure \ref{figure_pruned_parameter_count_vs_prunability} for an example of this problem and Section \ref{section_performance_of_prunability}) for our empirical results on this. 

While generalization measures have been studied for a long time \citep{vapnik2015uniform, mcallester1999pac, neyshabur2015normbased, zhang2016understanding, keskar2016large, nagarajan2019generalization, dziugaite2017computing}, \citet{jiang2019fantastic} recently brought new empirical rigor to the field. They perform a large-scale empirical study by generating a large set of trained neural networks with a wide range of generalization gaps. Proposing a range of new evaluation criteria, they test how predictive of a model's generalization performance previously proposed generalization measures actually are. In this paper, we evaluate our new generalization measure in the same framework and use the strongest measures as baselines for comparison against our proposed measure, the neural network's prunability. 

According to \citet{jiang2019fantastic}, three classes of generalization measures seem to be particularly predictive of generalization. First, measures that estimate the \textit{flatness} of local minima, either based on random perturbations of the weights \citep{neyshabur2017exploring, dziugaite2017computing, jiang2019fantastic, mcallester1999pac, arora2018stronger} or directly based on the curvature of the loss surface \citep{maddox2020rethinking}. In a purely empirical study, \citet{morcos2018importance} find that a model's robustness to clamping the \textit{activations} of individual units to fixed values is predictive of its generalization performance. Second, margin-based measures that use the distance of training samples to decision boundaries \citep{jiang2018predicting}. Lastly, \citet{jiang2019fantastic, ru2020revisiting} have proposed measures that estimate the speed of optimization.

We also evaluate our proposed generalization measure against another surprising phenomenon of generalization: Under certain conditions \textit{double descent} curves can be observed in which the test loss first decreases, then increases and then decreases again as a model's parameter count is increased \citep{he2016identity, nakkiran2019deep, belkin2018reconciling}. In contrast, many existing generalization measures grow monotonically with the model's size and thus fail to capture the double descent phenomenon, making it a particularly interesting setting to study new generalization measures \citep{maddox2020rethinking}.

Last, we provide a brief overview of recent research on neural network pruning. It is well established a large fraction of a network's parameters can be removed without impairing performance \citep{gale2019state, blalock2020state}; in fact, such pruning often \textit{improves} generalization performance \citep{thodberg1991improving}.  There are many approaches to selecting parameters for removal; we focus on randomly selected parameters (random pruning) and removing the smallest-magnitude parameters (magnitude pruning), though other approaches exist in the literature \citep{han2015learning, renda2020comparing, lee2018snip, Wang2020Picking, blalock2020state}. Random pruning is applied by the regularization method dropout \citep{srivastava2014dropout}, which is used to improve network generalization performance. Magnitude pruning can be used in an iterative pruning and retraining procedure \citep{frankle2018lottery} to identify sparse subnetworks that, for some sparsity levels, generalize better than their parent network. Both approaches are therefore natural candidates for a notion of network simplicity, which we will leverage in our proposed generalization measure.
In Dropout, random weights are temporarily removed during training to improve the models generalization performance at the end of training. In the Lottery Ticket research pruning is used to identify subnetworks which can be retrained to match the performance of the original network. In this paper, we study the distinct question of how we can use pruning after training and without any retraining to predict a model's generalization performance. In particular, we show that the larger the fraction of a model's parameters that can be removed without hurting the model's training loss, the better the model's generalization performance.

\vspacecaption
\subsection{Notation}
\label{section_notation}
Following the notation of \citet{jiang2019fantastic}, we denote by $\sD$ the distribution over inputs $\tX$ and corresponding labels $y$.
We write $\sS$ for a given data set containing $m$  tuples $\{(\tX_1, y_1),\dots, (\tX_m, y_m)\}$ drawn i.i.d. from $\sD$ where $\tX_i \in \mathcal{X}$ is the input data and $y_i \in \{1,\dots,\kappa\}$ the class labels. In that, $\kappa$ is the number of classes. Furthermore, we denote a feedforward neural network by $f_\vw: \mathcal{X} \rightarrow \mathbb{Z}$, its weight parameters by $\vw$ and the number of its parameters by $\omega$. For the weight tensor of the $i^{th}$ layer of the network we write $\tW_i$. The weight vector can be recovered as $\vw = vec(\tW_1, \dots, \tW_d)$, where $d$ is the depth of the network and $vec$ denotes the vectorization operator.

Let $L$ be the 1-0 classification error under the data distribution $\sD$ : $L(f_{\vw}) \defines \mathbb{E}_{(\tX, y) \sim \sD}[f_\vw(X) \not = y] $, and let $\hat L$ be the empirical 1-0 error on $\sS$: $\hat L(f_{\vw}) \defines \frac{1}{m}\sum_{i=1}^m [f_\vw(X_i) \not = y_i]$. We define the \textbf{generalization gap} as $g(f_\vw) \defines L(f_{\vw}) - \hat L(f_{\vw})$.
When training a network, we denote each hyperparameter by $\theta_i \in \Theta_i$ for $i = 1, \dots n$ where $n$ is the total number of hyperparameters.  We denote a particular hyperparameter configuration by $\vtheta \defines (\theta_1, \theta_2, \dots, \theta_n) \in \Theta$, where $\Theta \defines  \Theta_1 \times \Theta_2 \times \dots \times \Theta_n$ is the space of hyperparameters.

\vspacecaption
\section{Prunability}
\vspacecaptionlow
\label{section_prunability}
A generalization measure should satisfy three straightforward desiderata: 1) It should correlate with the generalization gap across a broad range of models. 2) Directly optimizing the measure during training should lead to improved generalization. 3) It provide new insights based on which theoretical understanding may be developed.

A measure based on magnitude- and random pruning presents a promising direction for satisfying all three desiderata. First, it is well-established that we can remove large fractions of a neural network's parameters without negatively impacting its train performance, while also improving its test performance \citep{blalock2020state, thodberg1991improving}. Second, training models using dropout, i.e. random pruning, is widely known to improve generalization \citep{srivastava2014dropout}.
Finally, pruning has a theoretical grounding in the minimum description length principle and in PAC-Bayesian bounds and further theoretical analysis might well tighten existing bounds.

\looseness -1 Indeed, we show how a simple extension of an existing PAC-Bayesian generalization bound \citep{mcallester1999pac} motivates the study of our proposed generalization measure. The PAC-Bayesian framework allows us to quantify the worst-case generalization performance of random predictors. PAC-Bayesian bounds define model complexity as the KL divergence between some fixed prior distribution $P$, and the probability distribution over functions induced by the random predictor (often called the posterior $Q$, although this may not correspond to a Bayesian posterior) to obtain generalization bounds of the following form.
\vspaceequation
\begin{theorem}\cite{mcallester1999pac}
For any $\delta > 0$, data distribution $\sD$, prior $P$, with probability $1 - \delta$ over the training set, for any posterior $Q$ the following bound holds:
$$\mathbb{E}_{\vv \sim Q} [L(f_\vv)] \leq \mathbb{E}_{\vv \sim Q} [\hat{L}(f_\vv)] + \sqrt{\frac{KL(Q||P) + \log(\frac{m}{\delta})}{2(m -1)}}$$
\end{theorem}
\citet{mcallester2013pacbayesian} shows that in the case where both the prior and posterior are distributions over networks to which dropout with a given dropout probability $\alpha$ is applied, the KL term becomes: $KL(Q||P) = \frac{1 - \alpha}{2}||\vw||^2$. We can directly use this formalization to derive a generalization bound for networks that are randomly pruned to the largest $\alpha$, s.t. $\mathbb{E}_{\vv \sim Q} [\hat{L}(f_\vv)] < \hat{L}(f_\vw) \times (1 + \beta)$, i.e., such that the training loss does not increase more than a relative tolerance $\beta$. We have to adjust the bound to account for the search for $\alpha$. In particular, we make use of the fact that we search over a fixed number $c$ of possible values for $\alpha \in [0, 1]$, using a union bound argument which changes the log term in the bound to $\log(\frac{cm}{\delta})$. As a result, we obtain the following bound on the expected generalization gap:
\vspaceequation
\begin{equation}\label{eq:pacbound}
\mathbb{E}_{\vv \sim Q}[g(f_v))] \leq \sqrt{\frac{\frac{1 - \alpha}{2}||\vw||_2^2 + \log(\frac{m}{\delta})  + \log(c)}{2(m -1)}}, \end{equation}
where $\mathbb{E}_{\vv \sim Q}[g(f_v))] = \mathbb{E}_{\vv \sim Q} [L(f_\vv)] - \mathbb{E}_{\vv \sim Q} [\hat{L}(f_\vv)]$.
While this bound will be vacuous for many large neural networks, it does provide us with some intuition for why we should expect a model's generalization to be linked to its robustness to pruning. In particular, this bound suggests that, the larger the fraction $\alpha$ of weights that can be pruned without hurting the model's training loss, the smaller the generalization gap $g(f_\vw)$ will be. It is this prediction that we empirically investigate in our experiments. Details on the derivation in can be found in Appendix \ref{section_pruning_pac_bayes_bound}.  
\vspacecaption
\subsection{Definition}
\vspacecaptionlow
Given a trained model and a pruning algorithm, we define the model's \textbf{prunability} as the smallest fraction of the model's parameters that we can keep while pruning without increasing the model's training loss by more than a given relative tolerance $\beta$. In Section \ref{section_experimental_results}, we show empirically that using the \textit{fraction} of remaining parameters as a measure of a model's complexity does indeed overcome the problems of using the absolute \textit{number} of parameters before or after pruning.
\vspaceequation
\begin{definition}[Prunability] Let $L$ denote the cross-entropy loss. For a given model $f_\vw$, a training data set $\sS_{train}$, a relative bound on the change in train loss $\beta$ and a pruning method $\phi$, we thus define prunability as
\vspaceequation
\begin{gather*}
\mu_{\textrm{prunability}}(f_\vw)  =  \argmin_\alpha  \phi(f_\vw, \alpha) \\
\textrm{s.t.} L(\phi(f_\vw, \alpha), \sS_{train}) \leq (1 + \beta) \times L(f_\vw, \sS_{train}) ~,
\end{gather*}

where $\phi(f_\vw, \alpha)$ sets the fraction $1-\alpha$ of the weights of $f_\vw$ to zero. 
\end{definition}
In our experiments we evaluate robustness to pruning both with regards to \textbf{magnitude pruning}, in which the parameters with the smallest absolute magnitude are set to 0, and \textbf{random pruning}, where random parameters are set to 0. Magnitude pruning can remove a greater number of model weights without affecting performance than random pruning \citep{blalock2020state}, and can therefore be considered a more effective pruning strategy. We hypothesize that using a more effective pruning method might yield a measure of prunability that is more predictive of generalization. Per default, we refer to a model's prunability with regards to magnitude pruning in the following. Note that while \textit{iterative} pruning is usually used when the aim is to achieve the best compression rate, we are only concerned with the generalization of the model at the end of training and thus apply one-shot pruning. In our experiments we use a tolerance $\beta$ of 0.1 and note that the particular choice of $\beta$ did not seem to have a large impact on the predictiveness of prunability. See Appendix \ref{section_prunability_algorithm} for pseudo-code of how prunability is computed for a given model.

While the above derivation serves as  motivation for our empirical study, further theoretical work will be necessary to fully explain the relation between a model's prunability and generalization. We see our empirical study as the central contribution of this paper and an important stepping stone towards a more comprehensive theoretical understanding of prunability.

\vspacecaption
\section{Experimental Results and Analysis}
\label{section_experimental_results}
\vspacecaptionlow

The two main goals of our experiments are to show that prunability is an effective generalization measure and that it measures something distinct from existing baselines. Our experiments aim to demonstrate this in four steps: 1) We verify that prunability predicts generalization across a set of models trained with different hyperparameters and generalization performance. To this end, we use a similar experimental framework as \citet{jiang2019fantastic}. 2) We evaluate how prunability performs in a particularly challenging double descent setting. 3) To understand what notion of simplicity prunability quantifies and whether prunability captures a different aspect of generalization than existing baselines, we study the conditional mutual information between it and baselines that represent existing classes of different strong generalization measures. 4) Given that prunability and random perturbation robustness both measure the sensitivity of a model's training loss to a perturbation of its weights, we directly compare how pruning and random weight perturbations impact models' losses.
\vspacecaption
\subsection{Baselines}
\vspacecaptionlow

That norm and parameter-count based generalization measures grow with model size is well established \citet{jiang2019fantastic} and the same applies for measures based on compressed models \cite{zhou2018non}. Instead, we compare our generalization measure to a number of strong baseline measures. Having somewhat tighter computational constraints than \citet{jiang2019fantastic}, we select one strong baseline per class of generalization measures as described by \citet{jiang2019fantastic} and \citet{jiang2018predicting}. 

From the class of generalization measures that aim to capture a model's simplicity using parameter-norms, we use the \textbf{Sum of Squared Frobenius Norms} \citep{jiang2019fantastic}: %
\vspaceequation
$$\mu_{\textrm{fro}}(f_\vw) = d\cdot(\prod_{i=1}^d||\tW_i||_F^2)^{1/d},$$ where $d$ is the number of layers of the model.

From the class of measures based on the idea that the speed of optimization is predictive of generalization, we use the negative \textbf{Sum of Train Losses} \citep{ru2020revisiting} throughout training:
\vspaceequation
$$\mu_{\textrm{SoTL}}(f_\vw) = - \sum_{t=1}^T \hat{L}(f_{w^t}),$$
where $T$ is the number of training epochs, $w^t$ the parameters at epoch $t$ and $\hat{L}$ the training cross-entropy loss.

 From the class of measures, based on the idea that models with flatter local minima generalize better, we use \textbf{magnitude aware random perturbation robustness} \citep{jiang2019fantastic}:
 \vspaceequation
 $$\mu_{\textrm{pac\_bayes\_mag\_flat}}(f_\vw) = \frac{1}{\sigma^2},$$
 where $u_i \sim \mathcal{N}(0, \sigma^2|w_i|+ \epsilon)$ and $\sigma^2$ is the largest number such that $\mathbb{E}_u[\hat{L}(f_{\vw + \vu})] \leq (1 + \beta) \times \hat{L}(f_{\vw}) $, where $\beta$ is a relative tolerance, in our experiments 0.1.
 
 Lastly, we evaluate a measure that was not tested by \citet{jiang2019fantastic} and which is much more complex generalization measure than the previously describe measures. This measure is part of the class of measures based on the idea that models with wider margins between the training samples and decision boundaries generalize better, we use the \textbf{Normalized Margins} measure \citep{jiang2018predicting}: 
 $$\mu_{\textrm{norm\_margins}}(f_\vw) = \va^T\vphi,$$
wherein $\vphi$ consists of five statistics of the normalized margins distribution for each layer of the model. 
That is, at each hidden layer of the network, the normalized distances of all samples to the decision boundary of the second most likely class are computed. Five statistics of each of these distance distributions are used as input for the linear regression. Note that this generalization measure is much more expensive to compute than the other measures under evaluation: it involves computing the margin for all training samples for multiple trained neural networks. Another limitation of this measure is that it is data set specific.
In Appendix \ref{appendix_generalization_measures} we provide further details on these generalization measures and additional baselines.
\vspacecaption
\subsection{Models}
\vspacecaptionlow
To generate a large set of trained models, we build upon the DEMOGEN set of trained models introduced by \cite{jiang2018predicting}. Our data set consists of 324 models that resemble the Network-in-Network architecture by \citep{lin2013network} trained on CIFAR10. \cite{jiang2018predicting} propose this particular architecture because it is parameter-efficient and achieves relatively competitive performance on standard image classification benchmarks. We achieve a wide range of generalization gaps by varying a number of hyperparameters such as the width, depth or $L_2$-regularization of the models. We modify the original set of models, DEMOGEN, used by \cite{jiang2018predicting} to obtain an even stronger experimental set up: noticing that the \textit{depth} of neural networks has a large impact on generalization performance, we add models of different depths to the original DEMOGEN data set. Following \citet{jiang2019fantastic} we only use models that apply batch normalization. See Appendix \ref{section_demogen_details} for further details of our experimental set-up.

\vspacecaption
\subsection{Evaluation Metrics}
\vspacecaptionlow
To study our generalization measures comprehensively, we consider 1) their Kendall's Rank correlation with the generalization gap, 2) how predictive they are in terms of the adjusted $R^2$ and 3) apply the conditional independence test of \citet{jiang2019fantastic}, in an attempt to understand the causal relations between the complexity measures and generalization. We briefly summarize the metrics in this section and provide a more detailed explanation in Appendix \ref{section_evaluation_metrics}

We compute \textbf{Kendall's rank correlation coefficient} which quantifies to what extent the ranking of models according to a given generalization measure corresponds to the ranking according to the generalization gaps. For a given generalization measure $\mu$, we consider: $\mathcal{T} \defines \cup_{\vtheta \in \Theta}\{(\mu(\vtheta), g(\vtheta))\}$ where $g(\vtheta)$ is the generalization gap of the model trained with hyperparameters $\vtheta$. Kendall's rank correlation coefficient is then defined as the fraction of pairs of tuples that are correctly ranked according to the generalization measure:
$\tau(\mathcal{T}) \defines \frac{1}{|\mathcal{T}|(|\mathcal{T}| - 1)} \sum_{(\mu_1, g_1) \in \mathcal{T} } \sum_{(\mu_2, g_2) \in \mathcal{T} \backslash (\mu_1, g_1)} \text{sign}(\mu_1 - \mu_2)\text{sign}(g_1 - g_2)$.
While Kendall's rank correlation coefficient is generally a useful metric, it does not adequately reflect whether a generalization measure's performance is consistent across all hyperparameter axes. Thus, we also consider the \textbf{granulated Kendall's rank correlation coefficient} which is essentially the average of Kendall’s rank coefficients $\psi_i$ obtained by only varying one hyperparameter dimension at a time: $\Psi \defines \frac{1}{n} \sum_{i=1}^n \psi_i$

To go beyond correlation, \citet{jiang2019fantastic} use a conditional independence test inspired by \citet{verma1991equivalence}. The main goal here is to understand whether there exists an edge in a causal graph between a given generalization measure $\mu$ and the generalization gap $g$: this tells us whether models generalize well \textit{because} of the value of the generalization measure, or whether the two values are only correlated. This is achieved by estimating the \textbf{conditional mutual information} between the generalization measure and the generalization gap conditioned on different hyperparameters that are observed. Formally, we have that for any function $\phi: \Theta \rightarrow \mathcal{R}$, let $V_\phi: \Theta_1 \times \Theta_2 \rightarrow \{+1, -1\}$ be as $V_\phi(\theta_1, \theta_2)  \defines sign(\phi(\theta_1) - \phi(\theta_2))$. Let $U_\mathcal{S}$ be a random variable that corresponds to the values of the hyperparameters in $\mathcal{S}$. Intuitively, the higher this metric is, the more likely it is that there is indeed an edge in the causal graph between the generalization gap and the generalization measure:  $\mathcal{K}(\mu) = \min_{U_\mathcal{S} s.t. |\mathcal{S}|\leq 2} \hat{\mathcal{I}}(V_\mu, V_g | U_\mathcal{S}),$ where $\hat{\mathcal{I}}(V_\mu, V_g | U_\mathcal{S})$ is the normalized conditional mutual information between the generalization measure and the generalization gap conditioned on observing the hyperparameter set $\mathcal{S}$.

\looseness -1 To evaluate how well we can predict a model's generalization using a linear function of a given generalization measure and following \cite{jiang2018predicting}, we additionally evaluate the generalization measures' predictiveness in terms of their adjusted $R^2$, i.e. the proportion of the variance of the generalization gap $g(\vtheta)$ across hyperparameters $\vtheta \in \Theta$ that can be explained by the generalization measure $\mu(\vtheta)$.

\vspacecaption
\subsection{Results}
\vspacecaptionlow
\subsubsection{Performance of Prunability on extended DEMOGEN data set}
\vspacecaptionlow
\label{section_performance_of_prunability}
Table \ref{table_generalization_measures_summary} holds a summary of our experimental results on the extended DEMOGEN data set. We find that prunability is highly informative of generalization across all of our evaluation metrics and that our measure is competitive with some of the strongest existing generalization measures from \citet{jiang2019fantastic}'s study of more than 40 generalization measures. In particular, it outperforms random perturbation robustness, the training loss itself and the Frobenius norm measures in terms of Kendall's rank correlation coefficient. The negative sum of training losses slightly outperforms prunability.
\begin{table}[tp]
\centering
\caption{\textbf{Prunability is competitive with strong baseline generalization measures.} Comparison of generalization measures' correlation with test performance on a set of convolutional networks trained on CIFAR-10. Higher values are better across all metrics. The standard error of the Kendall's $\tau$ is $s_{\tau} = 0.037$. \vspace{4pt}}
\label{table_generalization_measures_summary}
\resizebox{\columnwidth}{!}{%
\begin{tabular}{l l l l}
\toprule
\textbf{Generalization Measure} & \textbf{Kendall's $\bm{\tau}$} & \textbf{Adjusted R2} & \textbf{CMI} \\ \hline 
Low-complexity Measures & & & \\ \addlinespace
Prunability (ours)            & 0.6496 & 0.5440           & 0.0324                                 \\ 
Random Prunability (ours)                 & 0.584                & 0.3015               & 0.0377                                   \\ 
Random Perturbation Robustness  & 0.4253                 & 0.2440                & 0.0230                                  \\ 
Frobenius-norm                  & -0.6055                & \textbf{0.7778}               & \textbf{0.0612}                                 \\ 
Negative Sum of Train Losses          &\textbf{ 0.7085}                 & 0.4923    &     0.0186         \\

Train loss                      & 0.0201                &  0.0746             & 0.0004      \\
\hline
High-complexity Measures & & & \\
\addlinespace
Normalized Margins              & \textbf{0.8061}                 & \textbf{0.8866}   & \textbf{0.1761}             \\ 

Best Margin Variable            & 0.6344                & 0.6623    &   0.0743           \\ 
\bottomrule                         
\end{tabular}%
 }%
\vspace{-16pt}
\end{table}

In addition to the generalization measures studied by \citet{jiang2019fantastic}, we also compare prunability to the much more complex and computationally expensive Normalized Margins baseline. It is also the only of the measures we study which is based on the data space rather than the parameter space.  This baseline clearly outperforms prunability while the Best Margin Variable seem similarly informative of generalization as prunability. Note that the margin-based measures achieve much higher conditional mutual information scores than the other baselines, providing further evidence for the causal connection between wide margins and good generalization performance. The conditional mutual information between \textit{prunability} and generalization is relatively low, suggesting that it seems somewhat less likely that there is a direct causal connection between a model's prunability and its generalization performance. 

We find that while prunability based on magnitude pruning outperforms prunability based on random pruning, the difference in performance is relatively small.

In Table \ref{table_generalization_measures_detailed_kendall}, we compare the granulated and regular Kendall's coefficients of some additional baselines. The results reconfirm the weak performance of norm and parameter count-based complexity measures both for pruned and unpruned networks. While this a well known issue, we point out that this is also problematic for recently proposed generalization measures based on the size of the compressed network such as \citet{zhou2018non}. We also note that the Frobenius norm measure is strongly negatively correlated with generalization rather than positively correlated as one would expect based on \cite{jiang2019fantastic}. It seems that this result is largely due to a strong negative correlation for models with different dropout rates and relatively weak correlation coefficients along other hyperparameter axes.

\begin{table*}[h]
    \centering
     \caption{\textbf{Prunability achieves a stronger granulated rank correlation than measures based on random perturbation robustness or norms.} Columns labeled with hyperparameters (width, dropout rate, etc.) indicate Kendall $\tau$ if we only vary this hyperparameter. The last two columns are the Overall Kendall's $\tau$ and the Granulated Kendall's coefficient $\psi$, which is the average of the per-hyperparameter columns. Higher values are better across all columns \vspace{4pt}}
    \label{table_generalization_measures_detailed_kendall}
    \resizebox{\textwidth}{!}{%
    \begin{tabular}{l l l l  l l l l}%
\toprule    
\textbf{Generalization Measure}&\textbf{Width}&\textbf{Dropout Rate}&\textbf{Data Augment.}&\textbf{Weight Decay}&\textbf{Depth}&\textbf{Kendall $\tau$}&\textbf{$\mathbf{\vpsi}$}\\%
\hline \addlinespace%
prunability (ours)&\textbf{0.1582}&0.6986&0.0122&{-}0.001&0.2254&0.6496&0.2187\\%
random\_prunability (ours)&0.0714&0.6261&0.0464&0.0422&0.259&0.584&0.209\\%
random\_perturbation&0.1161&0.5154&0.0962&0.0755&0.1829&0.4253&0.1972\\%
frobenius\_norm&{-}0.0778&{-}0.7955&{-}0.067&{-}0.0591&0.4577&{-}0.6055&{-}0.1083\\%
normalized\_margins&0.0671&\textbf{0.8325}&\textbf{0.1121}&0.0799&\textbf{0.5789}&\textbf{0.8061}&0\textbf{.3341}\\%
best\_margins\_variable&0.054&0.7756&0.0415&{-}0.039&0.4798&0.6344&0.2624\\%
sum\_of\_train\_losses&-0.0639&0.8078&0.0498&0.0345&0.4156&0.7085&0.2487\\%
train\_loss&0.0143&{-}0.0652&{-}0.0762&{-}0.0349&0.1142&0.0201&{-}0.0096\\%
sum\_two\_norms&{-}0.0513&{-}0.5728&0.0214&0.0746&{-}0.1359&{-}0.2096&{-}0.1328\\%
parameter\_count&{-}0.0866&0.0&0.0&0.0&{-}0.4218&{-}0.0911&{-}0.1017\\%
sum\_two\_norms\_pruned&{-}0.0749&{-}0.6828&0.0857&\textbf{0.0866}&{-}0.2654&{-}0.2234&{-}0.1701\\%
pruned\_parameter\_count&{-}0.0893&0.6986&0.0122&{-}0.001&{-}0.3798&0.0417&0.0481\\%
\bottomrule
\end{tabular}%
}%
   
\end{table*}
\vspacecaption
\subsubsection{Double Descent}
\label{section_double_descent}
\vspacecaption
\label{section_double_descent_results}
\begin{figure}[]
    \centering
    %\subfloat[]{\label{sublable1}\includegraphics[width=0.9\columnwidth]{images/double_descent.pdf}} \\
    \subfloat[]{\label{sublable1}\includegraphics[width=0.9\columnwidth]{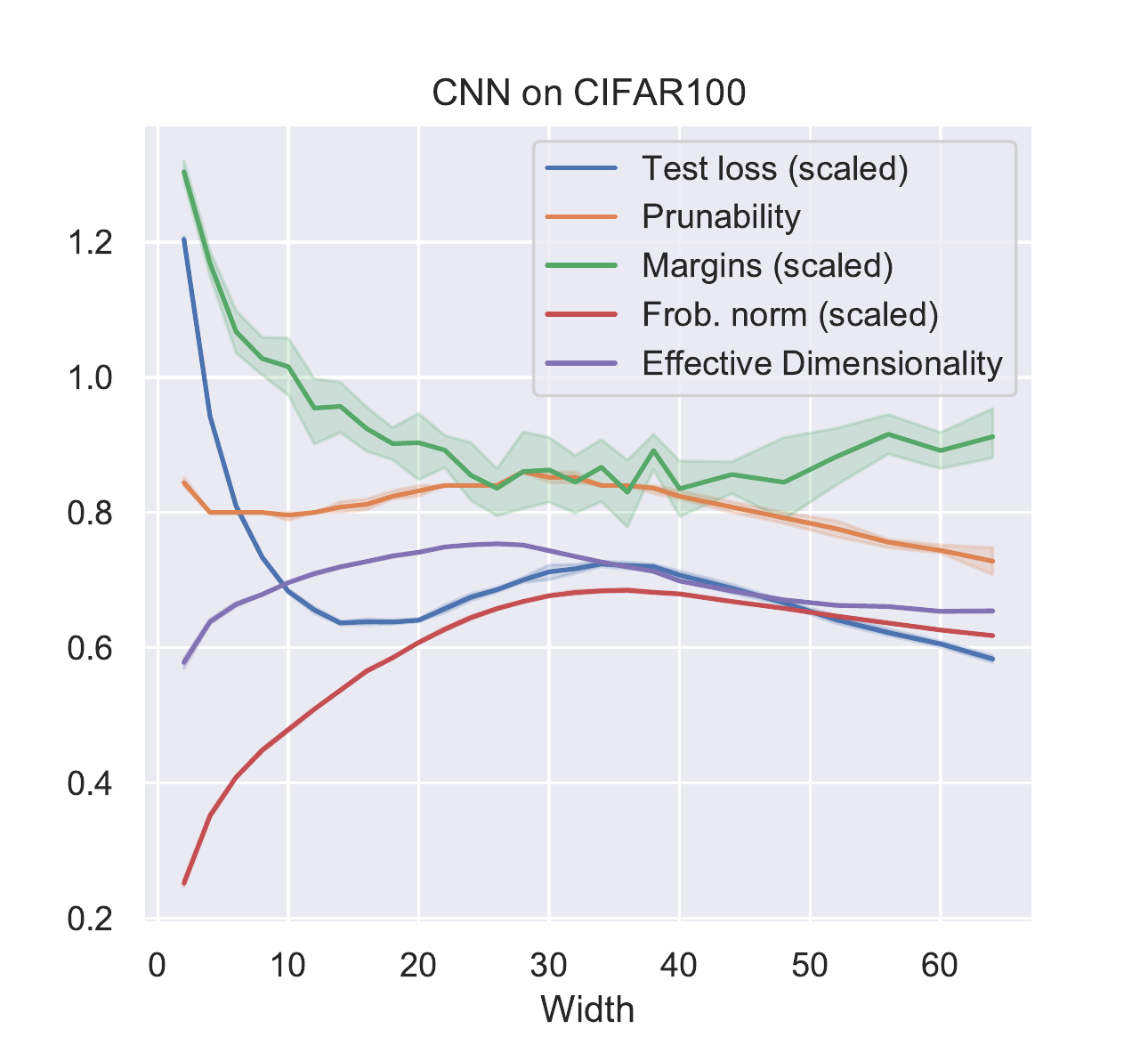}} \\
    \centering
    \subfloat[]{
\resizebox{0.8\linewidth}{!}{\begin{tabular}{l l l}
\toprule 
\textbf{Kendall's} $\tau$                         & \textbf{Test Loss} & \textbf{Test Error} \\ \hline \addlinespace
Prunability              & \textbf{0.3613 } & 0.1277       \\ 
Eff. Dim. &   -0.0197 & -0.0753       \\
Random Perturbation &   0.1398  & -0.2166       \\ 
Frobenius Norm &   0.0996 & -0.5367       \\ 
Margins &   0.1296 & \textbf{0.4258}       \\ 
-Sum of Train Losses &   -0.2890 & -0.9381       \\ 
\bottomrule
\end{tabular}}}
\caption{\textbf{Prunability is correlated with test loss in double descent setting:} Across a set of convolutional networks of varying width trained on CIFAR-100, we show that prunability captures double descent behavior better than a previously proposed metric \textit{Effective Dimensionality} – which is based on the eigenspectrum of the Hessian of the training loss of the model – and other strong baselines.} \label{tab:plot_double_descent}
\label{figure_double_descent}
\end{figure}%
In addition to our experiments on the DEMOGEN data set, we also evaluate prunability on a set of models that exhibit a test loss double descent \citep{nakkiran2019deep}. This phenomenon is particularly interesting for the study of generalization measures, since many measures that aim to capture a model's complexity grow monotonically as we increase a network's width and thus do not display any form of double descent. In the previous section we saw that \textit{prunability} seems highly predictive of generalization for models of different widths. The double descent setting allows us to test whether there is more to prunability's success than its correlation with model size. We use the experimental setup proposed by \citet{maddox2020rethinking}, and compare our new measure to their curvature-based generalization measure which also aims to capture double descent. Note that up to now we studied the generalization gap in terms of the $0-1$ loss. In this section, however, we focus on another aspect of generalization and study the relation between prunability and models' cross-entropy test loss.

\citet{maddox2020rethinking} introduce a generalization measure called Effective Dimensionality which is defined as:
$N_{\text{eff}}(\tH, z) = \sum_{i=1}^k \frac{\lambda_i}{\lambda_i + z}$ where $\lambda_i$ are the eigenvalues of the Hessian of the training loss $\tH \in \mathcal{R}^{k \times k}$, and $z > 0$ is a regularization constant. Intuitively, the Effective Dimensionality measures the flatness of the local minimum of a network in terms of the number of eigenvalues of the Hessian of the training loss that are "large". To get a measure on the same scale as our new generalization measure, we use a normalized version of $N_{eff}$: $\mu_{\textrm{effective\_dimensionality}} = \frac{N_\text{eff}}{k}$. Additionally, we also evaluate a number of the strongest baselines from Section \ref{section_performance_of_prunability} in the double descent setting.
We refer to Appendix \ref{section_double_descent_appendix} for further details on the experimental setup.

We find that prunability does indeed display a double descent and is more strongly rank correlated with the test loss than the baseline measures (Kendall's $\tau$ of 0.3613, $p < 0.0005$). See Figure \ref{figure_double_descent} for an overview of the results.

In Appendix \ref{section_double_descent_bonus_experiment} we study an additional double descent setting and obtain results that are consistent with the results presented in this section.

\vspacecaption
\subsubsection{Pruning and Flatness-based Measures}
\vspacecaption
\label{section_comparison_of_generalization_measures}
We now investigate whether a neural network's robustness to pruning can be explained by other established generalization measures, or whether it is measuring something entirely distinct from previous notions of model simplicity. To better understand the relation between prunability and the other generalization measures we perform the conditional independence test described in section \ref{section_conditional_independence} between them. Recall that, up to now, we used this test to study the relation between the generalization measure and the generalization gap and that the higher the $\kappa$ between two variables, the likelier it is that they neighbors in the causal DAG. See Table \ref{table_cmi_between_generalization_measures} for a summary of the results. 

First, we observe that the conditional mutual information between prunability and random perturbation robustness is the highest across the baselines we are evaluating. Because pruning is a special case of perturbation, some connection between the two methods is expected. However, prunability is clearly measuring something subtly different from flatness of the loss surface, as it outperforms random perturbation robustness in our evaluations. Thus, we investigate whether this distinction may arise from the differential effects of pruning and random perturbations in function space. 

\begin{figure}[h]
\begin{center}
%\framebox[4.0in]{$\;$}
\resizebox{1\columnwidth}{!}{\includegraphics{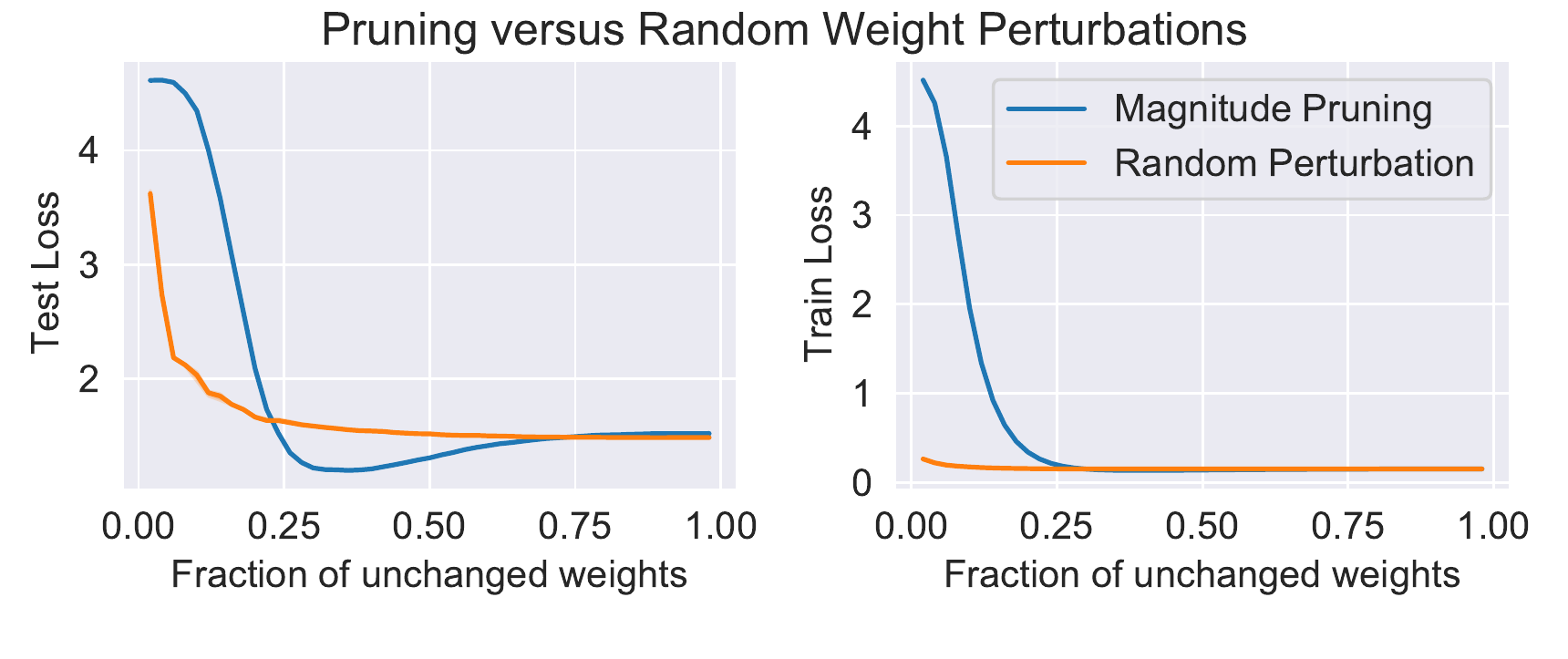}}
\end{center}\vspace{-16pt}
\caption{\textbf{Pruning affects models very differently than random perturbations.} Here we compare pruning of weights and randomly perturbing the same weights by the same amount. We study a ResNet18 trained on CIFAR100. Generally speaking, pruning will have a larger negative impact on a model's loss than randomly perturbing the same weights by the same amount but in some cases pruning actually improves the test loss of models.}
\label{figure_pruning_versus_random_perturbations}
\end{figure}

 To evaluate this, we directly compare the impact of both pruning and random perturbations on some of the models in our double descent experimental setting by performing pruning and random perturbations of the same magnitude and computing their respective impact on the loss. We find that pruning has a larger negative impact on a model's training loss than a random perturbation of the same magnitude (see Figure \ref{figure_pruning_versus_random_perturbations}). In contrast, pruning can lead to an improvement of the test loss which we do not observe for random perturbations; we conjecture that this relationship may be responsible for the difference between the two methods, though disentangling this phenomenon is left for future work. We also note that prunability achieves much higher Kendall's $\tau$, Adjusted $R^2$ and Conditional Mutual Information than random perturbation robustness.

In conclusion, pruning affects models differently from randomly perturbing their weights. Because pruning can in certain settings improve test performance (and hence seems to be more aligned with network simplicity) we suggest that this difference leads to the stronger performance of prunability as compared to random perturbation robustness.
We compare the impact of pruning and random perturbations on a model's training and test loss for an additional architecture in Appendix \ref{section_pruning_versus_random_perturbations_bonus_experiment}, and find that the results are consistent with the results presented in this section.
\begin{table}[h]
\centering
\resizebox{\columnwidth}{!}{%
\begin{tabular}{l l l l l l l}
\toprule
\textbf{}          & Margins     & Rand. Perturb. & Param. Count & Train Loss & F-Norm & SoTL \\ \hline \addlinespace

Prunability       & 0.0122             & \textbf{0.04541 }              & 0.0038          & 0.0004     & 0.0105  & 0.0013     \\
\bottomrule
\end{tabular}
}
\caption{\textbf{Prunability is similar to Random Perturbation Robustness.} The conditional mutual information (CMI) between prunability and random perturbation robustness.}
\label{table_cmi_between_generalization_measures}
\end{table}

%\vspace{-16pt}

\vspacecaption
\section{Conclusion}
\label{section_conclusion}
\vspacecaptionlow
In this paper, we have shown that a network's prunability, the smallest fraction of parameters we can keep while pruning without hurting the model's training loss, is highly informative of the network's generalization performance. Unlike many existing measures of model complexity, prunability correctly predicts that larger models generalize better than smaller models in contrast to many previous measures of model simplicity based on parameter counting and norms. Prunability further seems to capture something distinct from previously suggested measures based on robustness to random perturbations or the curvature of the loss landscape. 

\looseness -1 Empirical studies of surprising phenomena of generalization in deep neural networks have previously served as a valuable stepping stone towards more comprehensive theories of generalization \citep{nakkiran2019deep, frankle2018lottery}. Similarly, we believe that our empirical observations lay the foundation for further theoretical work to analyze the relationship between pruning and generalization. One direction is to build upon the PAC-Bayesian approach outlined in Section \ref{section_prunability}, e.g. similar to \citet{dziugaite2017computing}. 

\looseness -1 An additional direction of future work is to build on the mutual information experiments of Section~\ref{section_comparison_of_generalization_measures} to study the relationship between pruning and other generalization measures: for example, how the direction of pruning relates to the span of the main eigenvectors of the Hessian studied by \citet{maddox2020rethinking}. Given the conceptual similarity, it may also prove fruitful to study the relation between pruning \textit{after} training and pruning \textit{during} training, that is dropout, more closely.

\nocite{langley00}

\bibliography{prunability}

\begin{thebibliography}{37}
\providecommand{\natexlab}[1]{#1}
\providecommand{\url}[1]{\texttt{#1}}
\expandafter\ifx\csname urlstyle\endcsname\relax
  \providecommand{\doi}[1]{doi: #1}\else
  \providecommand{\doi}{doi: \begingroup \urlstyle{rm}\Url}\fi

\bibitem[Allen-Zhu et~al.(2019)Allen-Zhu, Li, and Liang]{allen2019learning}
Allen-Zhu, Z., Li, Y., and Liang, Y.
\newblock Learning and generalization in overparameterized neural networks,
  going beyond two layers.
\newblock In \emph{Advances in Neural Information Processing Systems}, pp.\
  6155--6166, 2019.
\newblock URL \url{https://arxiv.org/pdf/1811.04918.pdf}.

\bibitem[Arora et~al.(2018)Arora, Ge, Neyshabur, and Zhang]{arora2018stronger}
Arora, S., Ge, R., Neyshabur, B., and Zhang, Y.
\newblock Stronger generalization bounds for deep nets via a compression
  approach.
\newblock volume~80 of \emph{Proceedings of Machine Learning Research}, pp.\
  254--263, Stockholmsmässan, Stockholm Sweden, 10--15 Jul 2018. PMLR.
\newblock URL \url{http://proceedings.mlr.press/v80/arora18b.html}.

\bibitem[Bartlett et~al.(2017)Bartlett, Foster, and
  Telgarsky]{bartlett2017spectrally}
Bartlett, P.~L., Foster, D.~J., and Telgarsky, M.~J.
\newblock Spectrally-normalized margin bounds for neural networks.
\newblock In \emph{Advances in Neural Information Processing Systems}, pp.\
  6240--6249, 2017.
\newblock URL \url{https://arxiv.org/pdf/1706.08498.pdf}.

\bibitem[Belkin et~al.(2019)Belkin, Hsu, Ma, and Mandal]{belkin2018reconciling}
Belkin, M., Hsu, D., Ma, S., and Mandal, S.
\newblock Reconciling modern machine-learning practice and the classical
  bias–variance trade-off.
\newblock \emph{Proceedings of the National Academy of Sciences}, 116:\penalty0
  201903070, 07 2019.
\newblock \doi{10.1073/pnas.1903070116}.
\newblock URL \url{https://arxiv.org/pdf/1812.11118.pdf}.

\bibitem[Blalock et~al.(2020)Blalock, Ortiz, Frankle, and
  Guttag]{blalock2020state}
Blalock, D., Ortiz, J. J.~G., Frankle, J., and Guttag, J.
\newblock What is the state of neural network pruning?
\newblock In \emph{Proceedings of 3rd MLSys Conference}, 2020.
\newblock URL \url{https://arxiv.org/pdf/2003.03033.pdf}.

\bibitem[Dziugaite \& Roy(2017)Dziugaite and Roy]{dziugaite2017computing}
Dziugaite, G.~K. and Roy, D.~M.
\newblock Computing nonvacuous generalization bounds for deep (stochastic)
  neural networks with many more parameters than training data.
\newblock In \emph{Proceedings of the Thirty-Third Conference on Uncertainty in
  Artificial Intelligence}, 2017.
\newblock URL \url{https://arxiv.org/pdf/1703.11008.pdf}.

\bibitem[Frankle \& Carbin(2019)Frankle and Carbin]{frankle2018lottery}
Frankle, J. and Carbin, M.
\newblock The lottery ticket hypothesis: Finding sparse, trainable neural
  networks.
\newblock In \emph{International Conference on Learning Representations}, 2019.
\newblock URL \url{https://openreview.net/forum?id=rJl-b3RcF7}.

\bibitem[Gale et~al.(2019)Gale, Elsen, and Hooker]{gale2019state}
Gale, T., Elsen, E., and Hooker, S.
\newblock The state of sparsity in deep neural networks.
\newblock \emph{arXiv preprint arXiv:1902.09574}, 2019.
\newblock URL \url{https://arxiv.org/pdf/1902.09574.pdf}.

\bibitem[Han et~al.(2015)Han, Pool, Tran, and Dally]{han2015learning}
Han, S., Pool, J., Tran, J., and Dally, W.
\newblock Learning both weights and connections for efficient neural network.
\newblock In \emph{Advances in neural information processing systems}, pp.\
  1135--1143, 2015.
\newblock URL
  \url{https://papers.nips.cc/paper/2015/file/ae0eb3eed39d2bcef4622b2499a05fe6-Paper.pdf}.

\bibitem[He et~al.(2016)He, Zhang, Ren, and Sun]{he2016identity}
He, K., Zhang, X., Ren, S., and Sun, J.
\newblock Identity mappings in deep residual networks.
\newblock In \emph{European conference on computer vision}, pp.\  630--645.
  Springer, 2016.
\newblock URL \url{https://arxiv.org/pdf/1603.05027.pdf}.

\bibitem[Hinton \& Van~Camp(1993)Hinton and Van~Camp]{hinton1993keeping}
Hinton, G. and Van~Camp, D.
\newblock Keeping neural networks simple by minimizing the description length
  of the weights.
\newblock In \emph{in Proc. of the 6th Ann. ACM Conf. on Computational Learning
  Theory}. Citeseer, 1993.
\newblock URL \url{http://www.cs.toronto.edu/~fritz/absps/colt93.pdf}.

\bibitem[Izmailov et~al.(2018)Izmailov, Podoprikhin, Garipov, Vetrov, and
  Wilson]{izmailov2018averaging}
Izmailov, P., Podoprikhin, D., Garipov, T., Vetrov, D., and Wilson, A.~G.
\newblock Averaging weights leads to wider optima and better generalization.
\newblock In \emph{Proceedings of Conference on Uncertainty in Artificial
  Intelligence}, 2018.
\newblock URL \url{http://auai.org/uai2018/proceedings/papers/313.pdf}.

\bibitem[Jiang et~al.(2019)Jiang, Krishnan, Mobahi, and
  Bengio]{jiang2018predicting}
Jiang, Y., Krishnan, D., Mobahi, H., and Bengio, S.
\newblock Predicting the generalization gap in deep networks with margin
  distributions.
\newblock In \emph{International Conference on Learning Representations}, 2019.
\newblock URL \url{https://openreview.net/forum?id=HJlQfnCqKX}.

\bibitem[Jiang et~al.(2020)Jiang, Neyshabur, Mobahi, Krishnan, and
  Bengio]{jiang2019fantastic}
Jiang, Y., Neyshabur, B., Mobahi, H., Krishnan, D., and Bengio, S.
\newblock Fantastic generalization measures and where to find them.
\newblock In \emph{International Conference on Learning Representations}, 2020.
\newblock URL \url{https://openreview.net/forum?id=SJgIPJBFvH}.

\bibitem[Keskar et~al.(2016)Keskar, Mudigere, Nocedal, Smelyanskiy, and
  Tang]{keskar2016large}
Keskar, N.~S., Mudigere, D., Nocedal, J., Smelyanskiy, M., and Tang, P. T.~P.
\newblock On large-batch training for deep learning: Generalization gap and
  sharp minima.
\newblock \emph{arXiv preprint arXiv:1609.04836}, 2016.

\bibitem[Krizhevsky et~al.(2009)Krizhevsky, Nair, and
  Hinton]{krizhevsky2009cifar}
Krizhevsky, A., Nair, V., and Hinton, G.
\newblock Cifar-10 and cifar-100 datasets.
\newblock \emph{URl: https://www. cs. toronto. edu/kriz/cifar. html},
  6:\penalty0 1, 2009.

\bibitem[Lee et~al.(2019)Lee, Ajanthan, and Torr]{lee2018snip}
Lee, N., Ajanthan, T., and Torr, P.
\newblock {SNIP}: {SINGLE}-{SHOT} {NETWORK} {PRUNING} {BASED} {ON} {CONNECTION}
  {SENSITIVITY}.
\newblock In \emph{International Conference on Learning Representations}, 2019.
\newblock URL \url{https://openreview.net/forum?id=B1VZqjAcYX}.

\bibitem[Lin et~al.(2014)Lin, Chen, and Yan]{lin2013network}
Lin, M., Chen, Q., and Yan, S.
\newblock Network in network.
\newblock \emph{CoRR}, abs/1312.4400, 2014.
\newblock URL \url{https://arxiv.org/pdf/1312.4400.pdf}.

\bibitem[Maddox et~al.(2020)Maddox, Benton, and Wilson]{maddox2020rethinking}
Maddox, W.~J., Benton, G., and Wilson, A.~G.
\newblock Rethinking parameter counting in deep models: Effective
  dimensionality revisited, 2020.
\newblock URL \url{https://arxiv.org/pdf/2003.02139.pdf}.

\bibitem[McAllester(2013)]{mcallester2013pacbayesian}
McAllester, D.
\newblock A pac-bayesian tutorial with a dropout bound, 2013.
\newblock URL \url{https://arxiv.org/pdf/1307.2118.pdf}.

\bibitem[McAllester(1999)]{mcallester1999pac}
McAllester, D.~A.
\newblock Pac-bayesian model averaging.
\newblock In \emph{Proceedings of the twelfth annual conference on
  Computational learning theory}, pp.\  164--170, 1999.
\newblock URL \url{https://arxiv.org/pdf/1509.08864.pdf}.

\bibitem[Morcos et~al.(2018)Morcos, Barrett, Rabinowitz, and
  Botvinick]{morcos2018importance}
Morcos, A.~S., Barrett, D.~G., Rabinowitz, N.~C., and Botvinick, M.
\newblock On the importance of single directions for generalization.
\newblock In \emph{International Conference on Learning Representations}, 2018.
\newblock URL \url{https://arxiv.org/pdf/1803.06959.pdf}.

\bibitem[Nagarajan \& Kolter(2019)Nagarajan and
  Kolter]{nagarajan2019generalization}
Nagarajan, V. and Kolter, J.~Z.
\newblock Generalization in deep networks: The role of distance from
  initialization.
\newblock \emph{arXiv preprint arXiv:1901.01672}, 2019.

\bibitem[Nakkiran et~al.(2020)Nakkiran, Kaplun, Bansal, Yang, Barak, and
  Sutskever]{nakkiran2019deep}
Nakkiran, P., Kaplun, G., Bansal, Y., Yang, T., Barak, B., and Sutskever, I.
\newblock Deep double descent: Where bigger models and more data hurt.
\newblock In \emph{International Conference on Learning Representations}, 2020.
\newblock URL \url{https://openreview.net/forum?id=B1g5sA4twr}.

\bibitem[Neyshabur et~al.(2015)Neyshabur, Tomioka, and
  Srebro]{neyshabur2015normbased}
Neyshabur, B., Tomioka, R., and Srebro, N.
\newblock Norm-based capacity control in neural networks.
\newblock volume~40 of \emph{Proceedings of Machine Learning Research}, pp.\
  1376--1401, Paris, France, 03--06 Jul 2015. PMLR.
\newblock URL \url{http://proceedings.mlr.press/v40/Neyshabur15.html}.

\bibitem[Neyshabur et~al.(2017)Neyshabur, Bhojanapalli, McAllester, and
  Srebro]{neyshabur2017exploring}
Neyshabur, B., Bhojanapalli, S., McAllester, D., and Srebro, N.
\newblock Exploring generalization in deep learning.
\newblock In \emph{Advances in Neural Information Processing Systems}, pp.\
  5947--5956, 2017.
\newblock URL \url{https://arxiv.org/abs/1706.08947}.

\bibitem[Renda et~al.(2020)Renda, Frankle, and Carbin]{renda2020comparing}
Renda, A., Frankle, J., and Carbin, M.
\newblock Comparing rewinding and fine-tuning in neural network pruning.
\newblock In \emph{International Conference on Learning Representations}, 2020.
\newblock URL \url{https://openreview.net/forum?id=S1gSj0NKvB}.

\bibitem[Rissanen(1986)]{rissanen1986stochastic}
Rissanen, J.
\newblock Stochastic complexity and modeling.
\newblock \emph{The annals of statistics}, pp.\  1080--1100, 1986.
\newblock URL \url{https://projecteuclid.org/euclid.aos/1176350051}.

\bibitem[Ru et~al.(2020)Ru, Lyle, Schut, van~der Wilk, and
  Gal]{ru2020revisiting}
Ru, B., Lyle, C., Schut, L., van~der Wilk, M., and Gal, Y.
\newblock Revisiting the train loss: an efficient performance estimator for
  neural architecture search.
\newblock \emph{arXiv preprint arXiv:2006.04492}, 2020.
\newblock URL \url{https://arxiv.org/pdf/2006.04492.pdf}.

\bibitem[Srivastava et~al.(2014)Srivastava, Hinton, Krizhevsky, Sutskever, and
  Salakhutdinov]{srivastava2014dropout}
Srivastava, N., Hinton, G., Krizhevsky, A., Sutskever, I., and Salakhutdinov,
  R.
\newblock Dropout: a simple way to prevent neural networks from overfitting.
\newblock \emph{The journal of machine learning research}, 15\penalty0
  (1):\penalty0 1929--1958, 2014.
\newblock URL
  \url{https://jmlr.org/papers/volume15/srivastava14a/srivastava14a.pdf}.

\bibitem[Thodberg(1991)]{thodberg1991improving}
Thodberg, H.~H.
\newblock Improving generalization of neural networks through pruning.
\newblock \emph{International Journal of Neural Systems}, 1\penalty0
  (04):\penalty0 317--326, 1991.
\newblock URL
  \url{https://www.worldscientific.com/doi/10.1142/S0129065791000352}.

\bibitem[Vapnik \& Chervonenkis(1971)Vapnik and
  Chervonenkis]{vapnik2015uniform}
Vapnik, V.~N. and Chervonenkis, A.~Y.
\newblock On the uniform convergence of relative frequencies of events to their
  probabilities.
\newblock In \emph{Theory of probability and its applications}, pp.\  11--30.
  Springer, 1971.

\bibitem[Verma \& Pearl(1991)Verma and Pearl]{verma1991equivalence}
Verma, T. and Pearl, J.
\newblock \emph{Equivalence and synthesis of causal models}.
\newblock UCLA, Computer Science Department, 1991.
\newblock URL \url{https://arxiv.org/pdf/1304.1108.pdf}.

\bibitem[Wang et~al.(2020)Wang, Zhang, and Grosse]{Wang2020Picking}
Wang, C., Zhang, G., and Grosse, R.
\newblock Picking winning tickets before training by preserving gradient flow.
\newblock In \emph{International Conference on Learning Representations}, 2020.
\newblock URL \url{https://openreview.net/forum?id=SkgsACVKPH}.

\bibitem[Zhang et~al.(2016)Zhang, Bengio, Hardt, Recht, and
  Vinyals]{zhang2016understanding}
Zhang, C., Bengio, S., Hardt, M., Recht, B., and Vinyals, O.
\newblock Understanding deep learning requires rethinking generalization.
\newblock In \emph{International Conference on Learning Representations}, 2016.
\newblock URL \url{https://openreview.net/pdf?id=Sy8gdB9xx}.

\bibitem[Zhou et~al.(2019)Zhou, Veitch, Austern, Adams, and
  Orbanz]{zhou2018non}
Zhou, W., Veitch, V., Austern, M., Adams, R.~P., and Orbanz, P.
\newblock Non-vacuous generalization bounds at the imagenet scale: a
  pac-bayesian compression approach.
\newblock In \emph{International Conference on Learning Representations}, 2019.

\bibitem[Zhu \& Gupta(2018)Zhu and Gupta]{zhu2017prune}
Zhu, M.~H. and Gupta, S.
\newblock To prune, or not to prune: Exploring the efficacy of pruning for
  model compression.
\newblock In \emph{International Conference on Learning Representations
  (Workshop)}, 2018.
\newblock URL \url{https://openreview.net/forum?id=S1lN69AT-}.

\end{thebibliography}
\bibliographystyle{icml2021}

\clearpage
\appendix

\vspacecaption
\section{Experimental Setup}
\vspacecaptionlow
\subsection{Evaluation Metrics}
\vspacecaptionlow
\label{section_evaluation_metrics}
In this section we describe the metrics we use to evaluate the generalization measures. To study our generalization measures comprehensively, we consider the generalization measures' Kendall's Rank correlation with the generalization gap, how predictive they are in terms of Adjusted $R^2$ and \citet{jiang2019fantastic}.'s conditional independence test in an attempt to understand the causal relations between the complexity measures and generalization.
\vspacecaption
\subsubsection{Kendall's Rank Correlation Coefficient}
\vspacecaptionlow
This criterion evaluates the quality of a given generalization in terms of how well it \textit{ranks} models with different generalization gaps. Intuitively, a generalization measure ranks two models that were trained with hyperparameters $\vtheta_1$ and $\vtheta_2$  correctly iff $\mu(\vtheta_1) > \mu(\vtheta_2) \rightarrow g(\vtheta_1) > g(\vtheta_2)$ is satisified. Kendall's rank coefficient then measures for what fraction of models in our data set this holds. Kendall T takes values between 1 (perfect agreement) and -1 (rankings are exactly reversed) where a Kendall's t of 0 means that the complexity measure and generalization gap are independent.

Formally, we consider the set of tuples of generalization measures and generalization gaps:
    $$\mathcal{T} \defines \cup_{\vtheta \in \Theta}\{(\mu(\vtheta), g(\vtheta))\}$$
    
And Kendall's rank correlation coefficient is defined as:
\begin{equation*}
        \tau(\mathcal{T}) \defines \frac{1}{|\mathcal{T}|(|\mathcal{T}| - 1)} \sum_{i}^{|\mathcal{T}|} \sum_{\substack{j \\ j \neq i}}^{|\mathcal{T}|}  sign(\mu_i - \mu_j)sign(g_i - g_j)
\end{equation*}

\vspacecaption
\subsubsection{Granulated Kendall's Rank Coefficient}
\vspacecaptionlow
While Kendall's rank correlation coefficient is generally a useful metric, it does not adequately reflect whether a generalization measure's performance is consistent across all hyperparameter axes.
To account for this \citet{jiang2019fantastic} propose the granulated Kendall's rank-correlation coefficient which is essentially the average of Kendall's rank coefficients obtained by only varying one hyperparameter dimension at a time.

Formally, we have:

$$m_i \defines|\Theta_1 \times \dots \times \Theta_{i-1} \times \Theta_{i+1} \times ... \Theta_n|$$

and the Kendall's rank correlation coefficient is defined as:

$$\psi_i \defines \frac{1}{m_i}\sum_{\theta_1 \in \Theta_1} \dots \sum_{\theta_{i - 1} \in  \Theta_{i - 1}} \sum_{\theta_{i + 1} \in  \Theta_{i + 1}} \dots  \sum_{\theta_{n} \in  \Theta_{n}} \tau_i$$

where $\tau_i = \tau \large( \cup_{\theta_{i} \in  \Theta_{i}} (\mu(\vtheta), g(\vtheta)) \large)$.

Averaging all of these values, we obtain the granulated Kendall's rank correlation coefficient:

$$\Psi \defines \frac{1}{n} \sum_{i=1}^n \psi_i$$

Given these definitions, we obtain the following standard errors for the granulated Kendall's coefficients in our experiments. For the width, dropout rate, weight decay and depth measurements we hsve, $s_{\tau} = 0.00056$ and for the data augmentation measurements we have $s_\tau{0.000484}$. Lastly, for $\psi$, we have $s_{\tau} = 0.00011$

\vspacecaption
\subsubsection{Conditional Mutual Information}
\vspacecaptionlow
\label{section_conditional_independence}
To go beyond measures of correlation, \citet{jiang2019fantastic} use a conditional independence test inspired by \citet{verma1991equivalence}. The main goal here is to understand whether there exists an edge in a causal graph between a given generalization measure $\mu$ and the generalization gap $g$. This is achieved by estimating the mutual information between the generalization measure and the generalization gap conditioned on different hyperparameters being observed. The higher this metric is, the more likely it is that there is indeed an edge in the causal DAG between the generalization gap and the generalization measure. 

Formally, we have that for any function $\phi: \Theta \rightarrow \mathcal{R}$, let $V_\phi: \Theta_1 \times \Theta_2 \rightarrow \{+1, -1\}$ be as $V_\phi(\theta_1, \theta_2)  \defines sign(\phi(\theta_1) - \phi(\theta_2))$. Let $U_\mathcal{S}$ be a random variable that corresponds to the values of the hyperparameters in $\mathcal{S}$. The conditional mutual information can then be computed as: $$\mathcal{I}(V_{\mu}, V_g|U_{\mathcal{S}}) = \sum_{U_\mathcal{S}}p(U_\mathcal{S}) \sum_{V_\mu \in \{\pm 1\}} \sum_{V_g \in \{\pm 1\}} \beta(V_{\mu}, V_{\theta}| U_\mathcal{S})$$

where $$\beta(V_{\mu}, V_{\theta}| U_\mathcal{S}) \defines p(V_\mu, V_g | U_\mathcal{S})\log \Big(\frac{p(V_\mu, V_g|U_\mathcal{S})}{p(V_\mu | U_\mathcal{S})p(V_g|U_{\mathcal{S}})}\Big)$$
By dividing $\mathcal{I}(V_{\mu}, V_g|U_{\mathcal{S}})$ above by the conditional entropy of generalization:
$$\mathcal{H}(V_g|U_\mathcal{S}) = - \sum_{U_\mathcal{S}}p(U_{\mathcal{S}}) \sum_{V_\mu \in \{\pm 1\}} p(V_g | U_\mathcal{S})\log(p(V_g|U_\mathcal{S}))$$

we obtain the normalized criterion:
$$\hat{\mathcal{I}}(V_\mu, V_g | U_\mathcal{S}) = \frac{\mathcal{I}(V_\mu, V_g | U_\mathcal{S})}{\mathcal{H}(V_g|U_\mathcal{S})}$$

In the original Inductive Causation algorithm \citep{verma1991equivalence}, an edge in the causal graph is kept if there is no subset of hyperparameters $\mathcal{S}$ such that the two two nodes are independent, that is $\hat{\mathcal{I}}(V_\mu, V_g | U_\mathcal{S}) = 0$. Due to computational limitations, and  following \citet{jiang2019fantastic}, we only consider subsets of hyperparameters of size at most 2:
$$\mathcal{K}(\mu) = \min_{U_\mathcal{S} s.t. |\mathcal{S}|\leq 2} \hat{\mathcal{I}}(V_\mu, V_g | U_\mathcal{S})$$

Intuitively, the higher $\mathcal{K}(\mu)$, the more likely it is that there exists an edge between $\mu$ and generalization in the causal graph.
\vspacecaption
\subsubsection{Adjusted $R^2$}
\vspacecaptionlow
Lastly, we consider a metric to evaluate how predictive a generalization measure is of generalization based on a simple linear regression: $\hat{g} = \va^T\mu + b$. We find $\va$ and $b$ by minimizing the mean squared error: $a^*, b^* = \arg \min_{\va, b} \sum_i(\va^T\phi(\vmu_i) + b - g_i)^2$ 

On a test set of models we then compute the coefficient of determination $R^2$ which measures what fraction of the variance of the data ca be explained by the linear model:
$$R^2 = 1 - \frac{\sum_{j=1}^n(\hat{g}_j - g_j)^2}{\sum_{j=1}^n(g_j - \frac{1}{n}\sum_{j=1}^n g_j)^2}$$

We compute the $R^2$ using a 10-fold cross-validation.

Alternatively, one can use the adjusted $R^2$ score ($\bar{R}^2$)  which measures how well a linear model fits a training set and which accounts for the number of variables used in the regression:

$$\bar{R}^2 = 1 - (1 - R^2)\frac{n - 1}{n - dim(\vmu) - 1}$$
\vspacecaption
\subsection{Generalization Measures}
\label{appendix_generalization_measures}
\vspacecaptionlow
Additional generalization measures we consider are:
\begin{itemize}
        \item $\mu_{\textrm{best\_margins\_variable}}$: To account for the fact that the normalized margins variable is based on  a large number of input variables, we select the single independent variable that has the largest standardized coefficient in the Normalized Margins model and directly use it as a generalization measure.
    
    \item $\mu_{\textrm{train\_loss}}$: $\hat{L}(f_\vw)$, the cross entropy loss of the model $f_\vw$ on the training data set after training is finished.
    \item $\mu_{\textrm{parameter\_count}}$: The number of parameters $\omega$ of the model.
    \item $\mu_{\textrm{sum\_two\_norms}}$: $$\mu_{\textrm{sum\_of\_two\_norms}}(f_\vw) = (\sum_{i=1}^d||\tW_i||_2^2)$$
    \item $\mu_{\textrm{sum\_two\_norms\_pruned}}$: Essentially, the same as $\mu_{\textrm{sum\_two\_norms}}$, but applied to $f_\vw$ that we obtain when we compute $\mu_{prunability}$. Formally, we have: 
    \begin{gather*}
    \mu_{\textrm{sum\_two\_norms\_pruned}}(f_\vw)  =  \mu_{\textrm{sum\_two\_norms}}(\min_\alpha  \phi(f_\vw, \alpha))\\
    \textrm{s.t.} \hat{L}(\phi(f_\vw, \alpha), \sS_{train}) \leq (1 + \beta) \times \hat{L}(f_w, \sS_{train})   
    \end{gather*}

    \item Analogously, $\mu_{\textrm{pruned\_parameter\_count}}$ is the parameter count $\omega$ of the pruned model we obtain while computing $\mu_{\textrm{prunability}}$

\end{itemize}
\vspacecaption
\subsection{DEMOGEN}
\vspacecaptionlow
\label{section_demogen_details}
We extend the DEMOGEN data set in \citet{jiang2018predicting} with regards to two aspects. First, we follow \citet{jiang2019fantastic} and only consider models that are trained using batch normalization. Second, \citet{jiang2018predicting} do not consider networks of different depths. We consider this to be a crucial hyperparameter and a extend the original DEMOGEN data set by training additional models with different numbers of NiN-blocks than the models in the original data set.

We use the network architecture proposed by \citet{jiang2018predicting} in the DEMOGEN data set which is very similar to the architecture used in \citet{jiang2019fantastic}. The architecture is very similar to the Network in Network architecture in \citet{lin2013network} with the max pooling and dropout layer removed. We train the networks on CIFAR-10 \citep{krizhevsky2009cifar}.
The architecture we use is thus:
\begin{enumerate}
    \item The input layer
    \item 2, 3 or 4 NiN-blocks as described in Table \ref{table_nin_block}.
    \item A single convolution layer with kernel size 4x4 and stride 1.
\end{enumerate}

\begin{table}[]
\centering
\caption{A NiN block}
\label{table_nin_block}
\begin{tabular}{l l}
\toprule
\textbf{Layer Id }& \multicolumn{1}{c}{\textbf{Layer Type}} \\ \hline \addlinespace
0        & 3 x 3 convolution, stride 2     \\ 
1        & 1 x 1 convolution, stride 1     \\ 
2        & 1 x 1 convolution, stride 1      \\ 
\bottomrule
\end{tabular}
\end{table}

In this architecture we vary the following hyperparameters:
\begin{enumerate}
    \item Use channel sizes of 192, 288, and 384
    \item Apply dropout after NiN-blocks with $p = 0.0, 0.2, 0.5$
    \item Apply $l_2$ regularization with $\lambda = 0.0, 0.001, 0.005$.
    \item Train with and without data augumentation, that is random cropping, flipping and shifting.
    \item We consider two random initializations per hyperparameter configuration.
    \item Models with 2, 3 or 4 NiN blocks as described above.
\end{enumerate}

In total we are thus working with a data set of 324 trained models. We train the models with SGD with momentum ($\alpha = 0.9$) with a batch size of 128 and an initial learning rate of 0.01. The networks are trained for 380 epochs with a learning rate decay of 10x every 100 epochs. We base our implementation on the original code base: \url{https://github.com/google-research/google-research/tree/master/demogen}.

\vspacecaption
\subsection{Double Descent}
\vspacecaptionlow
\label{section_double_descent_appendix}
For our double descent experiments we use the models and baselines used by \citet{maddox2020rethinking} which are available on: \url{https://github.com/g-benton/hessian-eff-dim}. In particular, the CNN we use is the architecture used by \citet{nakkiran2019deep} which is availbable on: \url{https://gitlab.com/harvard-machine-learning/double-descent/-/blob/master/models/mcnn.py} Likewise, the ResNet18 architecture can be found here: \url{https://github.com/g-benton/hessian-eff-dim/blob/temp/hess/nets/resnet.py}.

The other baselines correspond to the ones from the DEMOGEN experiments described in Appendix \ref{section_demogen_details}. The margins  baseline used in this setting is based on the one used in \cite{jiang2018predicting}: a linear model is trained to predict the test loss based on the statistics of the margin distribution at the \textit{output} layer of a given model.

We train a set of 32 networks of widths 2, 4, 6, ..., 64 on CIFAR100 \citep{krizhevsky2009cifar} using SGD with a learning rate of $10^{-2}$,  momentum of 0.9, weight decay of $10^{-4}$ for 200 epochs with a batch size of 128. The learning rate decays to $10^{-4}$ on a piecewise constant learning rate schedule \citep{izmailov2018averaging}, beginning to decay on epoch 100. Random cropping and flipping are used for data augmentation which is turned off for the computation of eigenvalues. Our plots contain the results across 5 random seeds. The shaded areas indicate the standard error. All experiments were run on Nvidia GeForce RTX 2080 Ti GPUs.

\vspacecaption
\subsection{Random Perturbations versus Pruning}
\label{section_random_perutrbation_versus_prunin_experimental_set_up}
\vspacecaptionlow
We use the ResNet18 implementation from \url{https://gitlab.com/harvard-machine-learning/double-descent/-/blob/master/models/resnet18k.py} and use the same hyperparameters for training as in the double descent settings above. 
For every magnitude pruned model we also compute the randomly perturbed model in which we perturb the same set of weights by a random vector of the same size.
\vspacecaption
\subsection{Algorithms}
\vspacecaptionlow
\subsubsection{Algorithm to determine prunability}
\vspacecaptionlow
\label{section_prunability_algorithm}
\begin{algorithm}[]
   \caption{Maximal Lossless Magnitude Pruning}
   \label{unstructured_pruning_algorithm}
\begin{algorithmic}
   \STATE {\bfseries Input:} Original model $f_\vw$, train\_data, tolerance in change of loss $\beta$, step size for search step\_size
   \STATE {\bfseries Output:} Pruned Model $f_{\vw_{pruned}}$
   \STATE original\_train\_loss = $L$($f_w$, train\_data)
   \STATE fraction\_of\_weights\_to\_remove = 0.98\;
   \STATE $\vw_{pruned} = \textrm{prune}(\vw, fraction\_of\_weights\_to\_remove)$\;
   \WHILE{$L$($f_{\vw_{pruned}}$, train\_data) $> (1 + \beta) \times \mathrm{original\_train\_loss}$}
   \STATE $\vw_{pruned} = \textrm{prune}(\vw, fraction\_of\_weights\_to\_remove)$
   \STATE fraction\_of\_weights\_to\_remove -= step\_size\
   \ENDWHILE
\end{algorithmic}
\end{algorithm}

Where $\textrm{prune}(\vw, \textrm{sparisty})$ is an algorithm that sets \textit{sparsity} \% of $\vw$ to 0 and then returns the new vector. In our experiments we evaluate both \textit{magnitude pruning}, where the parameters with the smallest absolute magnitude are set to 0, and \textit{random pruning}, where random parameters are set to 0. 

Empirically, we find that the particular choice for the tolerance $\beta$ does not make a big difference with regards to predictiveness of prunability. In our experiments we use $\beta = 0.1$. Similarly, we consider $c=50$ possible values for the pruning fraction $\alpha$.

Note that to determine a model's prunability we use the differentiable cross-entropy loss rather than the 1-0 loss used elsewhere in the paper. 

\vspacecaption
\section{Additional Experiments}
\label{section_additional_experiments}
\vspacecaptionlow
\subsection{Double Descent}
\label{section_double_descent_bonus_experiment}
\vspacecaptionlow

\begin{figure}[t]
    \centering
     \subfloat[Comparison of prunability and baseline measures in a double descent setting. As indicated, some measures were scaled to fit onto the same y-axis. This Figure extends Figure \ref{figure_double_descent}. ]{\includegraphics[width=\columnwidth]{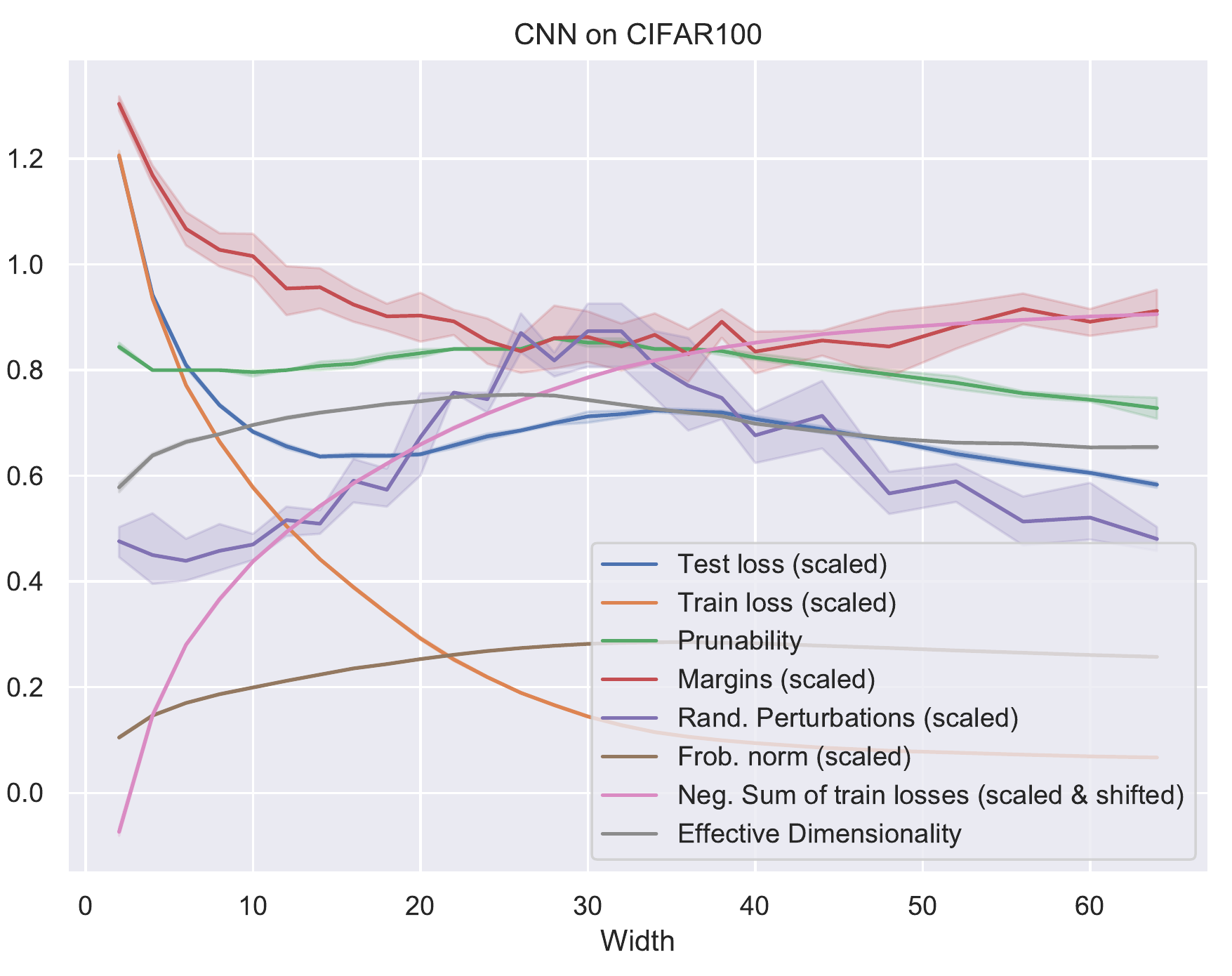}
          \label{fig:cnn_double_descent_bonus}}
         
     \hfill
     \centering
     \subfloat[Kendall's rank correlation ($s_{\tau}$ = 0.06) between generalization measures and test loss/test error. Higher values are better.]{

          \raisebox{-0.95in}{
          \begin{tabular}{l l l}
Kendall's $\tau$                         & Test Loss & Test Error \\ \hline  \addlinespace
Prunability              & \textbf{0.3613}  & 0.1277        \\ 
Eff. Dim. &   -0.0197 & -0.0753       \\ 
Random Perturbation &   0.1398  & -0.2166       \\ 
Frobenius Norm &   0.0996 & -0.5367       \\ 
Margins &   0.1296 & \textbf{0.4258}       \\ 
-Sum of Training Losses &   -0.2890 & -0.9381       \\ 
\end{tabular} }
        
         \label{tab:plot_double_descent}
     }
\caption{\textbf{Prunability is correlated with test loss in double descent setting:} Across a set of convolutional networks of varying width trained on CIFAR-100, we show that prunability captures double descent behavior better than a previously proposed metric \textit{Effective Dimensionality} – which is based on the eigenspectrum of the Hessian of the training loss of the model – and other strong baselines.} 
\label{figure_cnn_double_descent}
\end{figure}

\begin{figure}[t]
    \centering
    \vspace{2pt}
     \subfloat[Comparison of prunability and baseline measures in a double descent setting. As indicated, some measures were scaled to fit onto the same y-axis.]{\includegraphics[width=1\columnwidth, height=2.55in]{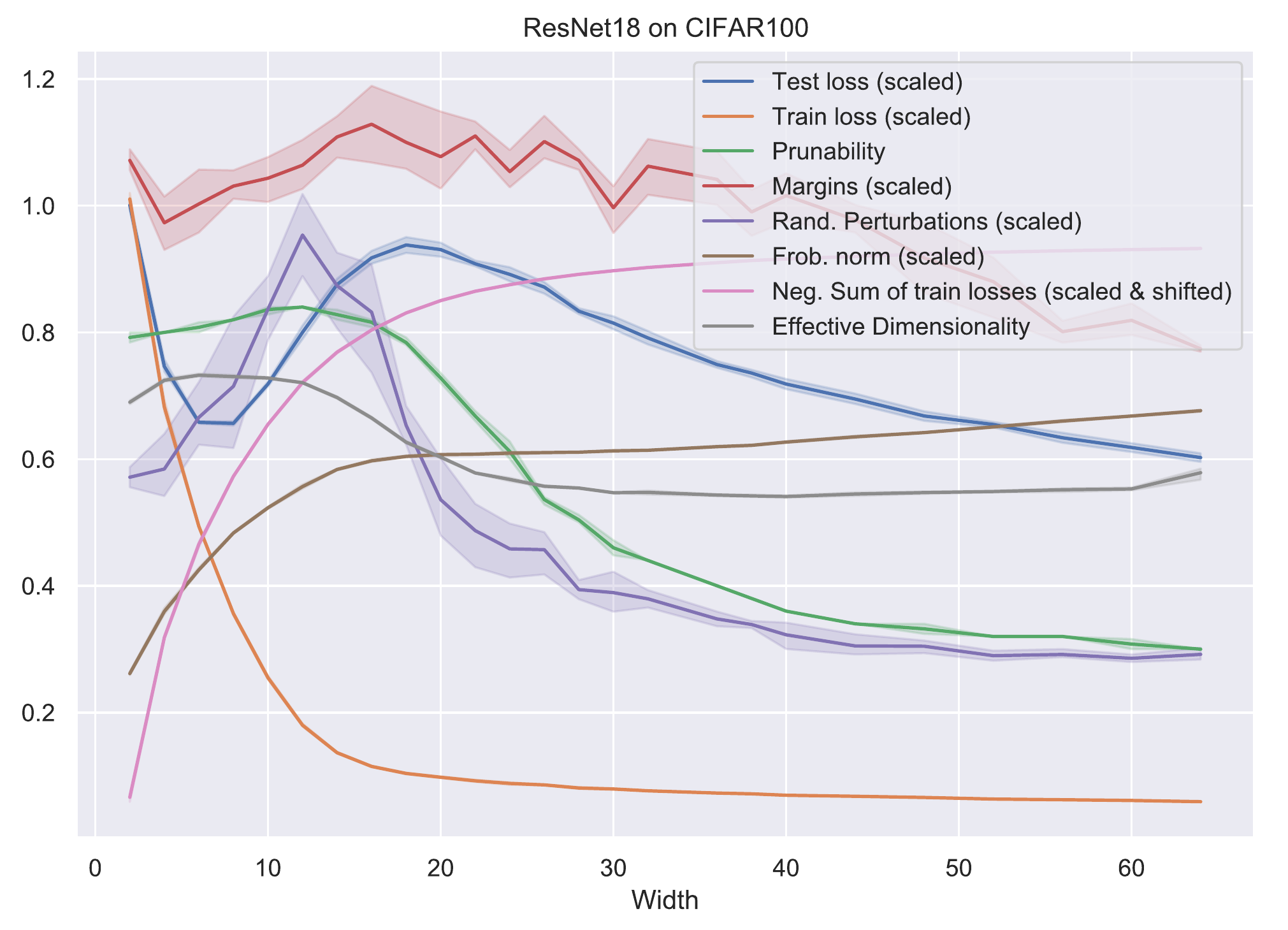}\label{fig:plot_double_descent}}
     \hfill
     \centering
     \subfloat[Kendall's rank correlation ($s_{\tau} = 0.06$) between generalization measures and test loss/test error. Higher values are better.]{

          \raisebox{-0.95in}{
          \begin{tabular}{l l l}
Kendall's $\tau$                         & Test Loss & Test Error \\ \hline \addlinespace
Prunability              & 0.4812  & \textbf{0.7556 }       \\ 
Eff. Dim. &   0.1272 & 0.5077       \\ 
Random Perturbation &   0.4532 & 0.7370       \\ 
Frobenius Norm &   -0.4487 & -0.9192       \\ 
Margins &   \textbf{0.5589} & 0.3883       \\ 
-Sum of Train Losses &   -0.4728 & -0.9393       \\ 
\end{tabular} }
 \label{tab:plot_double_descent}
     }
\caption{\textbf{Prunability is correlated with test loss in double descent setting:} Across a set of ResNet18s of varying width trained on CIFAR-100, we show that prunability captures double descent behavior better than a previously proposed metric \textit{Effective Dimensionality} – which is based on the eigenspectrum of the Hessian of the training loss of the model – and is competitive with other strong baselines.} 
\label{figure_resnet_double_descent}
\end{figure}

In Section \ref{section_double_descent}, we conduct an experiment that suggests that prunability captures the particularly challenging double descent phenomenon. In this section, we provide an extended version of Figure \ref{figure_double_descent}, containing all baselines (see Figure \ref{fig:cnn_double_descent_bonus}). Furthermore, we extend that experiment to an additional architecture, ResNet18 trained on CIFAR100 (see Figure \ref{figure_resnet_double_descent}). The results of this additional experiment are consistent with the one in the main part of the body: prunability is strongly Kendall rank correlated with the test loss and the test error. In this additional setting, however, the margins-based method outperforms prunability. We describe the details of the architecture and training of this experiment in Appendix \ref{section_random_perutrbation_versus_prunin_experimental_set_up}.

\subsection{Pruning versus Random Perturbations}
\label{section_pruning_versus_random_perturbations_bonus_experiment}
\begin{figure}[H]
\begin{center}
%\framebox[4.0in]{$\;$}
\resizebox{\columnwidth}{!}{\includegraphics{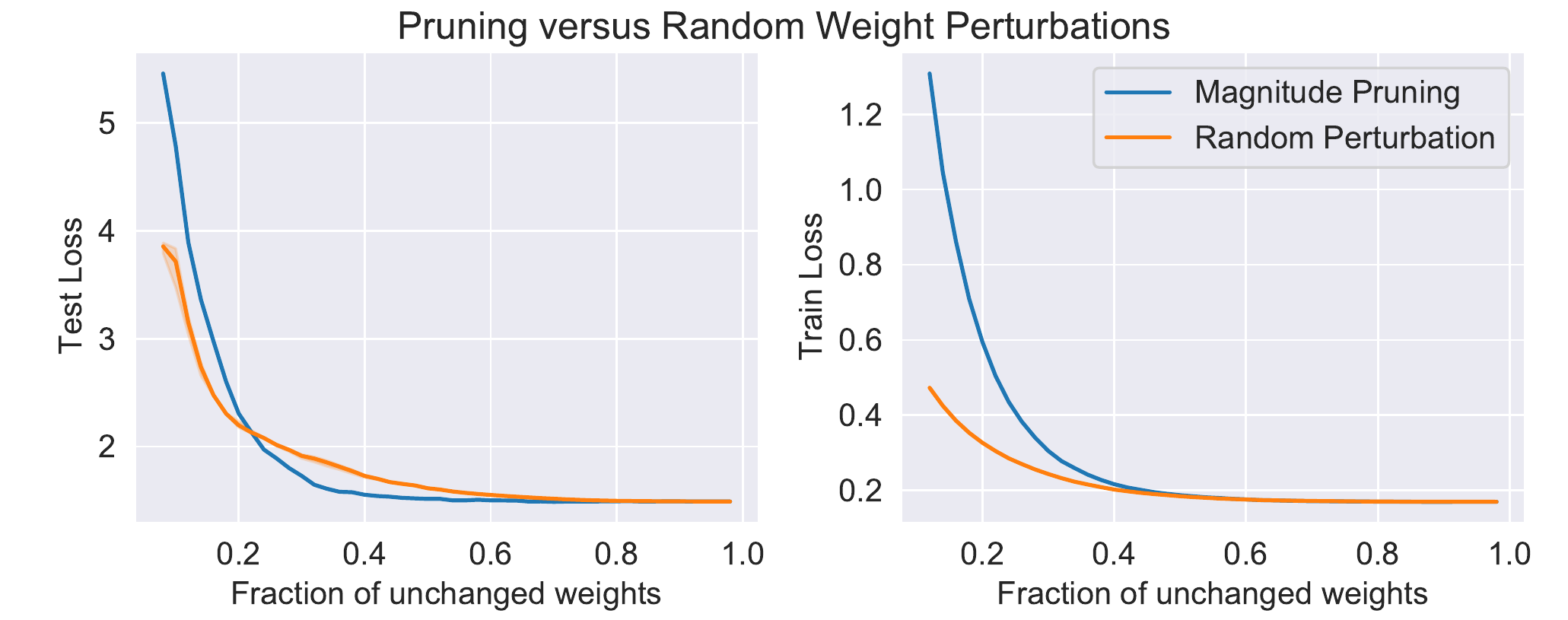}}
\end{center}
\caption{\textbf{Pruning affects models differently than random perturbations.} Here we compare pruning of weights and randomly perturbing the same weights by the same amount. We study a CNN on CIFAR100. Generally speaking, pruning will have a larger negative impact on a model's loss than randomly perturbing the same weights by the same amount. Similarly, as in the original experiments, we find that  pruning has a smaller negative impact on the test loss than randomly perturbing as long as a moderate number of parameters are being perturbed. For larger fractions of perturbed parameters, pruning has a larger negative impact on the test loss than random perturbations.}
\label{figure_convnet_pruning_versus_random_perturbations}
\end{figure}

Likewise, we study the impact of pruning and random perturbations on the training and test loss in an additional experimental setting. As in the experiment in Section \ref{section_comparison_of_generalization_measures}, we find that pruning and randomly perturbing weights (by a vector of the same magnitude), have very different impacts on the training and test loss (see Figure \ref{figure_convnet_pruning_versus_random_perturbations}). In this experiment, we study the CNN used in our double descent experiment of Section \ref{section_double_descent}. In particular, we find that pruning generally has a larger negative impact on the training loss than a random weight perturbation of the same magnitude. With regards to the test loss, we find that pruning has a smaller negative impact than random perturbations for moderate amounts of pruning, and a larger negative impact for larger amounts of pruning/perturbation. This matches the behavior of the first experiment. In this additional experiment we do not observe an improvement of the test loss through pruning which was not expected.

\vspacecaption
\section{PAC-Bayes bound}
\vspacecaptionlow
\label{section_pruning_pac_bayes_bound}

We use the following, simplified formulation of \cite{mcallester1999pac}'s dropout bounds which is also used in \cite{jiang2019fantastic}: \\

\textbf{Theorem 1} \textit{For any $\delta > 0$, distribution D, prior P, with probability $1 - \delta$ over the training set, for any posterior Q the following bound holds:}
\vspaceequation
$$\mathbb{E}_{\vv \sim Q} [L(f_\vv)] \leq \mathbb{E}_{\vv \sim Q} [\hat{L}(f_\vv)] + \sqrt{\frac{KL(Q||P) + \log(\frac{m}{\delta})}{2(m -1)}}$$
For the reader's convenience we briefly reproduce the derivation of the dropout bound in \citet{mcallester2013pacbayesian}.

To draw a sample $\omega$ from $Q$ means first drawing a sparsity pattern $S \sim S_\alpha$, where $\alpha$ is the dropout probability, and a noise vector $\epsilon \sim \mathcal{N}(0, 1)^d$. Then, we construct $\omega$ as $s \circ (\vw + \epsilon)$, $\circ$ being the Hadamard product: $(s \circ w)_i \defines  s_iw_i$. Similarly, a sample from $P$ is $s \circ \epsilon$.
 We denote by $S_\alpha$ the distribution on the $d$-cube $\mathcal{B}$ of sparsity patterns.
 We then have $\mathbb{E}_{\vv \sim Q} [f_\vv] = \mathbb{E}_{s \sim S_\alpha, \epsilon \sim \mathcal{N}(0,1)^d}[f_{s \circ (\vw + \epsilon)}]$.
We derive $KL(Q||P)$ as follows:

\begin{align*}
    KL(Q||P) &= \mathbb{E}_{s \sim S_\alpha, \epsilon \sim \mathcal{N}(0,1)^d}\Large{[}\ln\frac{S_\alpha(s) e^{-\frac{1}{2}||s\circ \epsilon||^2}}{S_\alpha(s) e^{-\frac{1}{2}||s\circ(\vw + \epsilon)||^2}}\Large{]} \\
    &= \mathbb{E}_{s \sim S_\alpha}[\frac{1}{2}||s \circ \vw||^2] \\
    &= \frac{1 - \alpha}{2}||\vw||^2
\end{align*}

We can directly obtain a bound for prunability from this, by searching for the largest $\alpha$, s.t.$\mathbb{E}_{\vv \sim Q} [\hat{L}(f_\vv)] < \hat{L}(f_\vw) \times (1 + \beta)$, where $\vw$ are the weights learned during training of the network.
We have to account for the fact that we search for $\alpha$ in our bound. We make use of the fact that we're searching over a fixed number $c$ of $\alpha$s and use union bound in the bound which will change the log term in the bound to $\log(\frac{cm}{\delta})$. 
Gives us the following bound: 
$$\mathbb{E}_{\vv \sim Q} [L(f_\vv)] \leq \mathbb{E}_{\vv \sim Q} [\hat{L}(f_\vv)] + \sqrt{\frac{\frac{1 - \alpha}{2}||\vw||_2^2 + \log(\frac{m}{\delta})  + 5}{2(m -1)}}$$

%%%%%%%%%%%%%%%%%%%%%%%%%%%%%%%%%%%%%%%%%%%%%%%%%%%%%%%%%%%%%%%%%%%%%%%%%%%%%%%
%%%%%%%%%%%%%%%%%%%%%%%%%%%%%%%%%%%%%%%%%%%%%%%%%%%%%%%%%%%%%%%%%%%%%%%%%%%%%%%

\end{document}